\documentclass[journal]{IEEEtran}
\usepackage{graphicx}
\usepackage{multirow}
\usepackage{longtable}
\usepackage{rotating}
\usepackage{color}
\usepackage{makecell}
\usepackage{hyperref}
\usepackage{cite}
\usepackage{lineno}
\usepackage{amsfonts}
\usepackage{amsmath}
\usepackage{stfloats}
\usepackage{booktabs}
\usepackage[table,xcdraw]{xcolor}
\usepackage{algorithm}
\usepackage{algorithmicx}
\usepackage{algpseudocode}
\usepackage{float}
\graphicspath{{figures/}}

\ifCLASSINFOpdf
\else
\fi

\begin{document}

\title{A Deep Decomposition Network \\for Image Processing: \\A Case Study of Visible and Infrared Image Fusion}
%

\markboth{Journal of \LaTeX\ Class Files,~Vol.~14, No.~8, August~2015}%
{Shell \MakeLowercase{\textit{et al.}}: Bare Demo of IEEEtran.cls for IEEE Journals}

\author{Yu~Fu,
	    Tianyang~Xu,
        Xiao-Jun~Wu,
\thanks{Y. Fu and X.-J. Wu are with the School of Artificial Intelligence and Computer Science, Jiangnan University, Wuxi 214122, P.R. China. (e-mail: yu\_fu\_stu@outlook.com,  wu\_xiaojun@jiangnan.edu.cn)}
\thanks{T. Xu is with the School of Artificial Intelligence and Computer Science, Jiangnan University, Wuxi 214122, P.R. China and the Centre for Vision, Speech and Signal Processing, University of Surrey, Guildford, GU2 7XH, UK. (e-mail: tianyang\_xu@163.com)}
}

\maketitle

\begin{abstract}
Image decomposition into constituent components has many applications in the field of image processing. It aims to extract salient features from the source image for subsequent pattern recognition. In this paper, we propose a new image decomposition method based on a convolutional neural network. It is applied to the task of image fusion. In particular, a pair of infrared and visible light images are decomposed into three high-frequency feature images and a low-frequency feature images respectively. The two sets of feature images are fused using a novel fusion strategy to obtain fused feature maps. The feature maps are subsequently reconstructed to obtain the fused image. Compared with the state-of-the-art fusion methods, the proposed approach achieves better performance in both subjective and objective evaluation.
\end{abstract}


\IEEEpeerreviewmaketitle

\section{Introduction}
\IEEEPARstart{I}{mage} fusion is an important task in image processing. 
It aims to extract important features from images of multi-modal sources and reconstruct the fused image using the complementary information conveyed by the multiple pictures by means of a fusion strategy. 
In general, there are several concrete image fusion tasks, involving the visible light and infrared image fusion, multi-exposure image fusion, medical image fusion and multi-focus Image Fusion. 
At present, diverse fusion methods are used in a range of applications, such as security surveillance, autonomous driving, medical diagnosis, camera photography, target recognition and other fields, for this purpose.
The existing fusion methods can be divided into two categories: traditional methods and deep learning methods \cite{liu2018deep}. 
Most of the traditional approaches apply signal processing methods to decompose the multi-modal images into high-frequency and low-frequency signals and then merge them.
With the development of deep learning, methods based on deep neural networks have also shown a great potential in image fusion.

Traditional methods can further be subdivided  into two categories: i) multi-scale decomposition approaches, and ii) representation learning methods. 
In the multi-scale domain, the image is decomposed into multi-scale representation feature maps. 
The multi-scale feature representations are then fused by various specific fusion strategies. 
Finally, the corresponding inverse transform is used to obtain the fused image. 
There are many representative multi-scale decomposition methods, such as pyramid \cite{mertens2009exposure}, curvelet \cite{zhang1999categorization}, contourlet \cite{upla2014edge},discrete wavelet transform \cite{hamza2005a}, etc.

In the representation learning domain. 
Most methods are based on sparse representation such as sparse representation (SR) with gradient histogram (HOG) \cite{zong2017medical}, joint sparse representation (JSR) \cite{zhang2013dictionary}, approximate sparse representation with multi-selection strategy \cite{bin2016efficient}.
In the low-rank domain, Li and Wu et al. proposed a low-rank representation (LRR)  based fusion method \cite{li2017multi}. 
In addition, recent advanced approaches, such as MDLatLRR \cite{li2020mdlatlrr} are based on image decomposition with Latent LRR. 
This method can extract source image features in the low-rank domains.

Although the methods based on multi-scale decomposition and representation learning have achieved promising performance, they still have some problems. 
These methods are very complicated, and the dictionary learning stage is a time-consuming operation especially for online training. 
Besides, if the source image is complex, these methods will not be able to extract the features well.
In order to solve this problem, in recent years, many methods based on deep learning have been proposed \cite{liu2018deep} to take advantage of the powerful feature extraction capabilities of neural networks.

In 2017, Liu et al. proposed a method based on convolutional neural network for multi-focus image fusion \cite{liu2017multi-focus}. 
In ICCV2017, Prabhakar et al. proposed DeepFuse \cite{prabhakar2017deepfuse} to solve the problem of multi-exposure image fusion. 
In 2018, Li and Wu et al. proposed a new paradigm for infrared and visible light image fusion, based on denseblock and autoencoder structure \cite{li2018densefuse}. 
In the subsequent two years, with the rapid development of deep learning, a large number of excellent methods emerged. 
They include IFCNN \cite{zhang2020ifcnn} proposed by Zhang et al.,  the fusion network based on GANs (PerceptionGAN) \cite{fu2021image} proposed by Fu et al., the multi-scale fusion network framework (NestFuse) \cite{li2020nestfuse} proposed by Li et al. (2020), the Transformer-based fusion network(PPTFusion) \cite{fu2021ppt} and DualFusion \cite{fu2021dual} based on Dual-branch autoencoder. 
Most of these methods use the powerful feature extraction function of neural networks, and perform fusion at the feature level. 
The fused image is then obtained by some specific decoding strategies.

However, the methods based on deep networks also have some shortcomings: 1) As a feature extraction tool, neural network cannot explain the meaning of the extracted features. 
2) The network is complex and takes a long time to train. 
3) The available size of multimodal paired datasets is small, and many methods resort to using other data sets for training. 
This is not ideal for extracting multimodal images.

To solve these problems, we draw on both, the traditional and deep learning methods, and propose a novel network that can be used to decompose images. 
Taking infrared and visible light image fusion as an example, the proposed network decomposes infrared and visible light images into high and low frequency signals. 
Novel fusion rules are then used to combine the decomposed images, and the fused low and high frequency signals recombined to reconstruct the output fused image.  
The key innovation of the proposed method is that it makes use of deep neural network for both, image decomposition, as well as feature extraction. 
Compared with the state-of-the-art methods, our fusion framework achieves better performance in both subjective and objective evaluation.

This paper is structured as follows. 
In Section \ref{section:RELATED}, we introduce the closest related work. 
In Section \ref{section:PROPOSED}, we described our proposed fusion method in detail.
In Section \ref{section:EXPERIMENTS}, we introduce the experiments carried on to validate our method and discuss the experimental settings. 
The experimental results are analyzed and compared to baseline methods in the same section. 
Finally, in the last section \ref{section:CONCLUSIONS}, we draw the paper to conclusion.

\section{RELATED WORKS}
\label{section:RELATED}
There are many effective methods, whether based on traditional image signal processing or deep learning. 
Both frameworks inspired the work presented in this paper.

\subsection{Wavelet Decomposition and Laplacian Filter}
\begin{figure}[!ht]
	\centering
	\includegraphics[width=8cm]{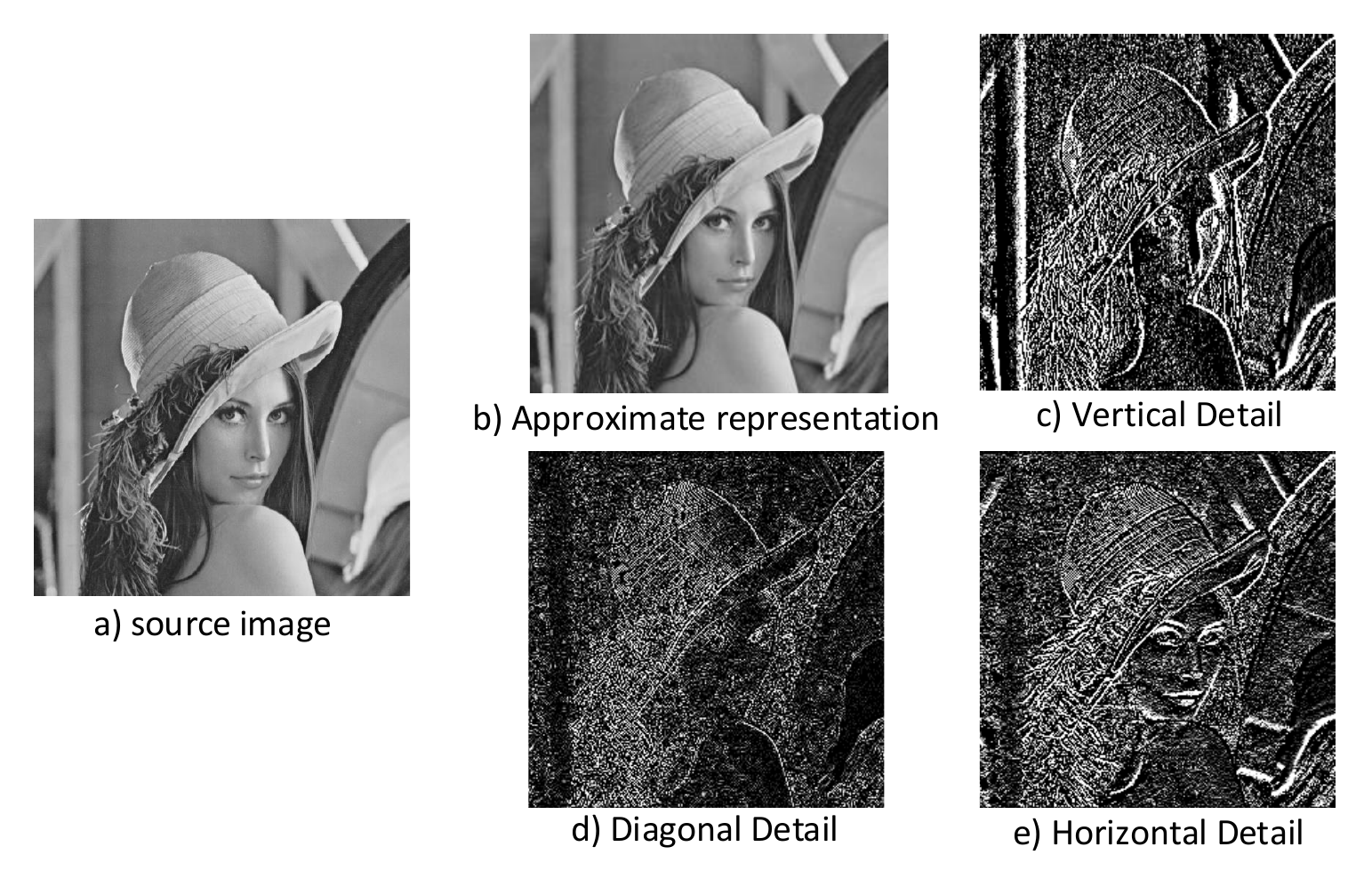}
	\caption{Wavelet decomposition produces a low-frequency image (b), and three high-frequency images (c)(d)(e) in the three spatial directions.}
	\label{fig:wavelet}
	\vspace{-6pt}
\end{figure}

The wavelet transform has been successfully applied to many image processing tasks.
The most common wavelet transform technique for image fusion is the Discrete Wavelet Transform (DWT) \cite{li1995multisensor,chipman1995wavelets}. 
DWT is a signal processing tool that decomposes signals into high and low frequency signals. 
Generally speaking, low-frequency information contains the main characteristics of the signal, whereas high-frequency information conveys the details. 
In the field of image processing, 2-D DWT is usually used to decompose images. 
The wavelet decomposition of the image is given as follows:
\begin{equation}
\begin{array}{l}
M_{L L}(x, y)=\phi(x) \phi(y) \\
M_{L H}(x, y)=\phi(x) \psi(y) \\
M_{H L}(x, y)=\psi(x) \phi(y) \\
M_{H H}(x, y)=\psi(x) \psi(y)
\end{array}\end{equation}
where $\phi(\cdot)$ is a low-pass filter, and $\psi(\cdot)$ is a high-pass filter.
The input signal $M_(x, y)$ is an image combining the signals of all bands. 
Along the $x$ and $y$ directions, high-pass and low-pass filtering are performed respectively. 
As shown in Fig.\ref{fig:wavelet}, the low-frequency image is a coarse approximation of the content of the three high-frequency images which contain the vertical detail, diagonal detail, and horizontal detail, respectively.

In addition, the Laplacian operator is a simple differential operator having the rotation invariance properties. 
The Laplacian transform of a two-dimensional image function is the isotropic second derivative, defined as:
\begin{equation}
\begin{aligned}
&\nabla^2f(x,y)=\frac{\partial^2 f(x,y)}{\partial x^2}+
\frac{\partial^2 f(x,y)}{\partial y^2}
\end{aligned}
\end{equation}
For digital image processing the equation is approximated by its discrete form as:

\begin{equation}
\vspace{-6pt}
\begin{aligned}
&\nabla^2f(x,y)\\
&\approx \![f(x\!+\!1,y)\! +\! f(x\!-\!1,y)\!+\!f(x,y\!+\!1)f(x,y\!-\!1)]\!-\!4f(x,y)
\end{aligned}
\vspace{-6pt}
\end{equation}

\begin{figure}[!ht]
	\centering
	\includegraphics[width=8cm]{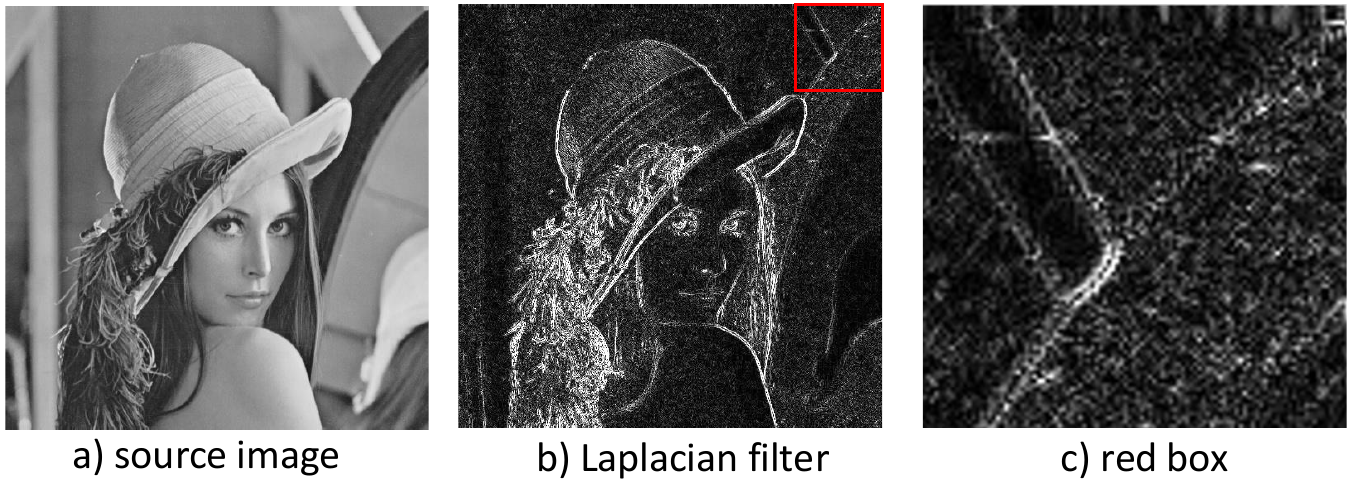}
	\caption{Laplacian Filter is used to extract the high-frequency image content (b) from the source image (a). The amplified detail of a local area of the high-frequency image is shown in (c).}
	\label{fig:laplacian}
	\vspace{-6pt}
\end{figure}

The Laplacian operator can also be expressed in the form of a convolution template, using it as a filtering kernel:

\begin{equation}
\vspace{-6pt}
\label{equ:mc}
G_1={\left[ \begin{array}{ccc}
	0 & 1 & 0\\
	1 & -4 & 1\\
	0 & 1 & 0
	\end{array} 
	\right ]},
G_2={\left[ \begin{array}{ccc}
	1 & 1 & 1\\
	1& -8 & 1\\
	1 & 1 & 1
	\end{array}
	\right ]}
\end{equation}
where $G_1$ and $G_2$ are the template and the extended template of the discrete Laplacian operator. 
The second differential characteristic of this template is used to determine the position of the edge. 
The filters are often used in image edge detection and image sharpening, as shown in Fig.\ref{fig:laplacian}.

It should be noted that the traditional edge filtering operator is just a high-frequency filter. 
While highlighting the edges, the filters also amplify the noise.

\subsection{Decomposition-based Fusion Methods}
\begin{figure}[!ht]
	\centering
	\includegraphics[width=\linewidth]{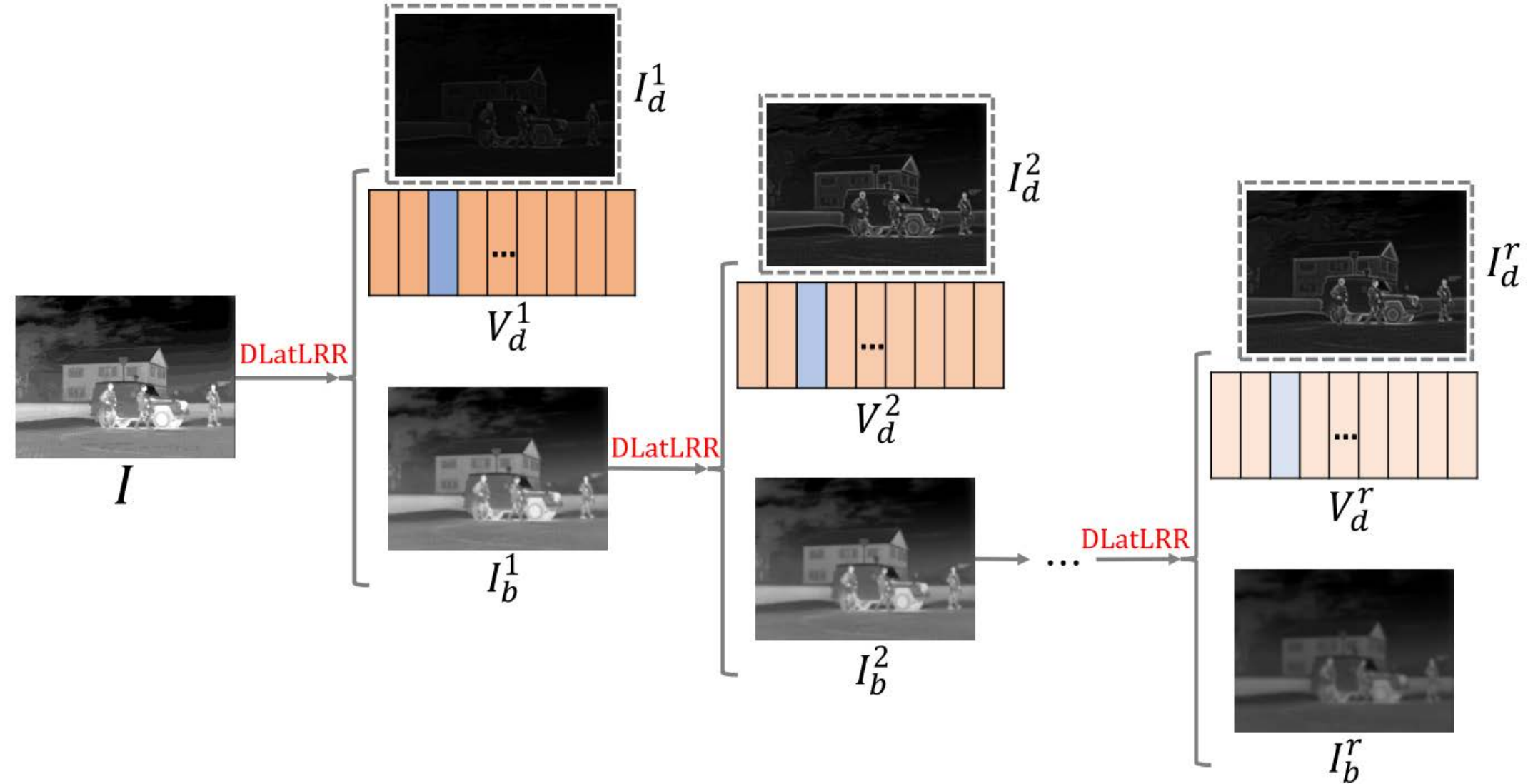}
	\caption{The framework of MDLatLRR.}
	\label{fig:MDLatLRR}
	\vspace{-6pt}
\end{figure}

Li and Wu et al. proposed a method \cite{li2020mdlatlrr} to decompose images using low-rank representation \cite{liu2011latent} (LatLRR).
In essence, their method can be described by the following optimization problem:
\begin{equation}\begin{array}{l}
\min _{Z, L, E}\|Z\|_{*}+\|L\|_{*}+\mu\|E\|_{1} \\
\text {s.t.}, X=X Z+L X+E
\end{array}\end{equation}
where $\mu$ is a hyper-parameter, $\|\cdot\|_{*}$ is the nuclear norm, and $\|\cdot\|_{1}$ is the $l_1$ norm. 
$X$ is observed data matrix. 
$Z$ is the low-rank coefficients matrix. 
$L$ is a projection matrix. 
$E$ is a sparse noisy matrix.
The authors use this method to decompose the image into detail image $I_{d}$ and base image $I_{b}$. 
We can see from Fig .\ref{fig:MDLatLRR} that $I_{d}$ is a high-frequency image, and $I_{b}$ is a low-frequency image.
 
%

As shown in Fig .\ref{fig:MDLatLRR}, the low-frequency image $I_{b}$ is further decomposed to obtain a cascade of high-frequency images $I_{d1}$, $I_{d2}$ and $I_{d3}$.

Similarly, this method decomposes the infrared image and the visible light image to obtain high-frequency images and low-frequency images. 
After fusing the information in the decomposed frequency, the fused image $I_{f}$ is obtained by means of reconstruction.

\subsection{Deep Learning-based Fusion Methods}
In 2017, Liu et al. proposed a neural network-based method \cite{liu2017multi-focus}.
The authors divide the image into many small patches. 
Then CNN is used to determine whether each small patch is blurry or clear. 
The network builds a decision activation map to indicate which pixels of the original image are clear and which are focused. 
The authors showed that a well-trained network can accomplish multi-focus fusion tasks very well, but their method is not suitable for other image fusion applications.

In order to enable the network to fuse visible light images and infrared images, Li and Wu proposed a deep neural network (DenseFuse) \cite{li2018densefuse} based on an auto-encoder. 
First they train a powerful encoder and decoder to extract the features of the original image, and then reconstruct it without losing information. 
The infrared image and the visible light image are encoded.  
The two sets of coding features are then fused. 
Finally, the fused features are input into the decoder to obtain the fused image.
These methods use the encoder to decompose the image into several latent features.
Then these features are fused and reconstructed to obtain the fused image.

In the past few years, Generative Adversarial Networks (GANs) have also been applied to image fusion. 
The pioneering work based on this approach is FusionGan \cite{ma2019fusiongan}. 
The generator inputs infrared and visible light images and outputs a fused image. 
In order to improve the quality of the generated image, the authors designed an application-specific loss function. 

These deep learning-based methods have the ability to map images into a high-dimensional feature space to process and fuse features. 
CNN can effectively extract the different dimensions images features. 
However, the CNN-based image fusion methods needs to a well-designed network structure, and can only be applied to specific image fusion tasks.
In view of the evident effectiveness of image decomposition and neural network based feature extraction,  we propose a novel method which combines multi-layer image decomposition and feature extraction using a unified neural network framework for infrared and visible light image fusion.

\section{PROPOSED FUSION METHOD}
\label{section:PROPOSED}
In this section, the proposed multi-scale decomposition-based fusion network is introduced in detail. 
Firstly, the fusion framework is presented in section \ref{subsection:Network}. 
Then, the details of the training phase are described in Section \ref{subsection:Training}. 
Next, in Section \ref{subsection:Loss} we design the loss function for the network training. 
Finally, we present different fusion strategies in Section \ref{subsection:Fusion}. 

\subsection{Network Structure}
\label{subsection:Training}
\begin{figure*}[!ht]
	\centering\resizebox{13cm}{!}{
	\includegraphics[width=\linewidth]{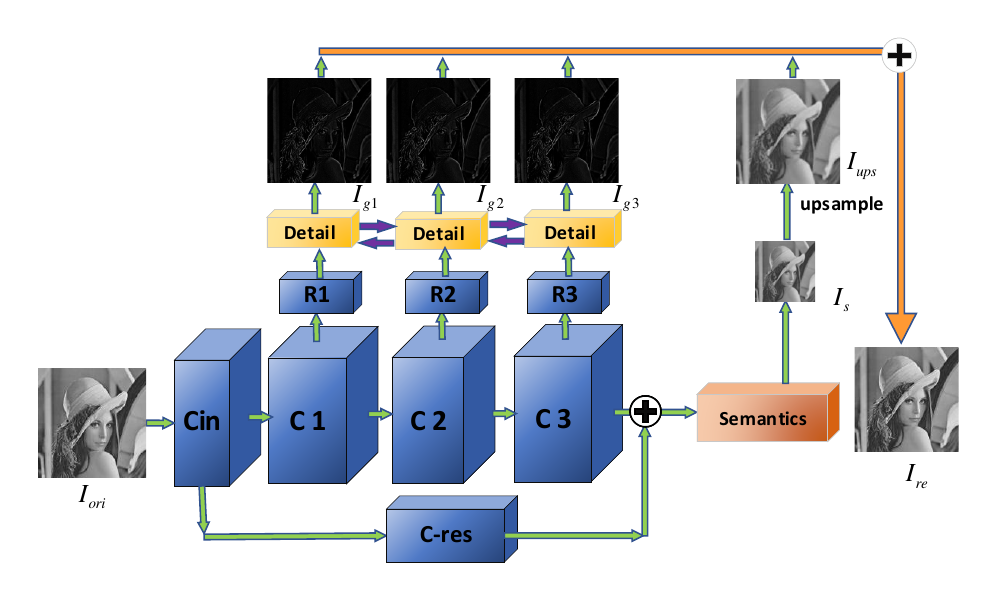}
	}
	\caption{The framework of the training process. }
	\label{fig:backbone}
	\vspace{-6pt}
\end{figure*}

Our goal is to construct and train a deep neural network so that it decomposes the source image into several high-frequency signals and one low-frequency signal for subsequent fusion operations. 
The structure of the network is shown in Fig. \ref{fig:backbone}, and the network settings are detailed in Table \ref{tab:parameters}.

\begin{figure}[!ht]
	\centering
	\includegraphics[width=6cm]{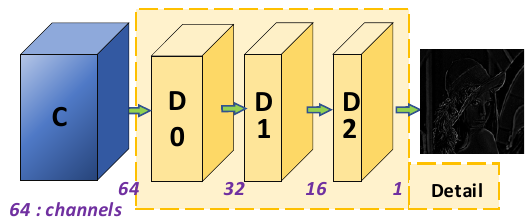}
	\caption{We use three convolutions to reduce the channels to one and output a high-frequency image.}
	\label{fig:detail}
	\vspace{-6pt}
\end{figure}

\begin{figure}[!ht]
	\centering
	\includegraphics[width=6cm]{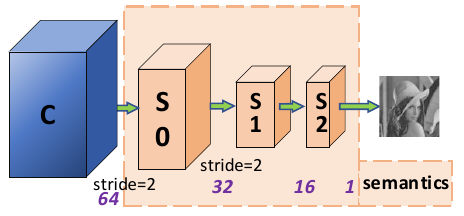}
	\caption{We use three convolutions and downsampling twice to reduce the channels to one and produce a low-frequency image.}
	\label{fig:semantic}
	\vspace{-6pt}
\end{figure}

\begin{table}[!ht]
\vspace{-6pt}
	\centering
	\caption{\label{tab:parameters}The parameters of the network  }
	\resizebox{\linewidth}{!}{
		\begin{tabular}{@{}cccccccc@{}}
			\toprule
			\multirow{2}{*}{Block}    & \multirow{2}{*}{Layer} & Channel & Channel  & Size     & Size    & Size     & \multirow{2}{*}{Activation} \\
			&                        & (input) & (output) & (kernel) & (input) & (output) &                             \\ \midrule
			\multirow{3}{*}{Cin}      & Conv(Cin-1)            & 1       & 16       & 3        & 256     & 256      & LeakyReLU                   \\
			& Conv(Cin-2)            & 16      & 32       & 3        & 256     & 256      & LeakyReLU                   \\
			& Conv(Cin-3)            & 32      & 64       & 3        & 256     & 256      & LeakyReLU                   \\ \midrule
			C1                        & Conv(C1)               & 64      & 64       & 3        & 256     & 256      & LeakyReLU                   \\
			C2                        & Conv(C2)               & 64      & 64       & 3        & 256     & 256      & LeakyReLU                   \\
			C3                        & Conv(C3)               & 64      & 64       & 3        & 256     & 256      & LeakyReLU                   \\ \midrule
			R1                        & Conv(R1)               & 64      & 64       & 1        & 256     & 256      & -                           \\
			R2                        & Conv(R2)               & 64      & 64       & 1        & 256     & 256      & -                           \\
			R3                        & Conv(R3)               & 64      & 64       & 1        & 256     & 256      & -                           \\ \midrule
			\multirow{3}{*}{Detail}   & Conv(D0)               & 64      & 32       & 3        & 256     & 256      & LeakyReLU                        \\
			& Conv(D1)               & 32      & 16       & 3        & 256     & 256      & LeakyReLU                        \\
			& Conv(D2)               & 16      & 1        & 3        & 256     & 256      & Tanh                        \\ \midrule
			\multirow{3}{*}{C-res}    & Conv(C-res1)           & 64      & 64       & 3        & 256     & 256      & ReLU                        \\
			& Conv(C-res2)           & 64      & 64       & 3        & 256     & 256      & ReLU                        \\
			& Conv(C-res3)           & 64      & 64       & 3        & 256     & 256      & ReLU                        \\ \midrule
			\multirow{3}{*}{Semantic} & Conv(S0)               & 64      & 32       & 3        & 256     & 128      & ReLU                        \\
			& Conv(S1)               & 32      & 16       & 3        & 128     & 64       & ReLU                        \\
			& Conv(S2)               & 16      & 1        & 3        & 64      & 64       & Tanh                        \\ \midrule
			Upsample                  & Upsample               & 1       & 1        & -        & 64      & 256      & -                           \\ \bottomrule
		\end{tabular}
	}
\end{table}

In Fig. \ref{fig:backbone} and Table \ref{tab:parameters}, $I_{ori}$ is the original input image, and $I_{re}$ is the reconstructed image. 
The backbone of the network is constituted by four feature extraction convolutional blocks ($Cin, C1, C2, C3$). 
The low-frequency feature extraction part is the $'semantic'$ block in the figure. 
The $'semantic'$ block shown in Fig. \ref{fig:semantic} includes two down-sampling convolutional layers ($S0, S1$) with a stride of 2 and a common convolutional layer ($S3$) which generates a low-resolution semantic image $I_{s}$. 
Then $I_{s}$ is up-sampled to the same size as $I_{ori}$ to obtain the low-frequency image $I_{ups}$. 

We copy the features of different depth ($C1, C2,C3$) and and then reshuffle their channels with convolutional layers ($R1, R2, R3$). 
Subsequently, we input them into the $'detail'$ branch of the shared weights to obtain three high-frequency images $I_{g1}$, $I_{g2}$ and $I_{g3}$. 
The detail branch here is shown in Fig. \ref{fig:detail} and Table \ref{tab:parameters}, which includes three convolutions (D0, D1, D2). 
The number of channels is reduced to 1 to obtain a high-frequency image. 
The reason for adding reshuffle layers ($R1, R2, R3$) here is that the detail block is weight-sharing. 
The feature maps that extract high-frequency information should follow the same channel distribution. 
So we add a $1\times1$ convolutional layer that does not share weights, and reshuffle and sort the feature channels so that the features can adapt to the weight-shared details block.

Finally, the three high-frequency images ($I_{g1}$, $I_{g2}$, $I_{g3}$) and one low-frequency image ($I_{ups}$) are added pixel by pixel to obtain the final reconstructed image $I_{re}$. 
Note that the high-frequency image and the low-frequency image are complementary. 
In fact, when the network learns to generate images, the high-frequency image can be viewed as the residual data of the low-frequency image. 
Bearing this in mind, we reflect this reality in terms of a residual branch ($'C-res'$ block). 
We skip-connect the result of $cin$ directly to the front of the $semantic$ block, adding it to the result of $c3$, and input it to the following layers. 
In this way, $C1, C2$ and $C3$ process the residual data between the source image and the semantic image intuitively. 
In order to make the skip-connected data more closely match the deep features of $C3$, we performed three convolutions in $'C-res'$ block to enhance the semantics of the skip-connected features.

As shown in the activation function in Table \ref{tab:parameters}, to consider the  properties of low-frequency and high-frequency images jointly, we choose LeakyRelu \cite{maas2013rectifier} as the activation function of the convolution layers in the backbone network and the high frequency components ($Cin, C1, C2, C3, detail$), and the Relu function is used as the activation function of the convolution layers of the residual branch ($'C-res'$) and the $semantic$ block ($S0,S1,S2$). 
The output of Relu has a certain degree of sparseness, which allows our low-frequency features to filter out irrelevant information and retain more blurred but semantic information. 
As a final consideration, in order to constrain the pixel value of the resulting image to a controllable range, we use the $Tanh$ activation function at the last layer of the $detail$ and $semantic$ blocks.

In general, the convolution block performs a convolution operation ($cin$) to obtain a set of feature maps containing various features. 
After the three identical convolution operations($c1,c2,c3$), three sets of shallow features are extracted. 
Applying two downsampling ($semantic$) operations, a deep feature is obtained. 
We argue that shallow features contain more low-level information such as texture and detailed features. 
We reshuffle the channels and feed these three sets of shallow features into the high-frequency branch $(detail$) to obtain three high-frequency images.
As deep features, in contrast, contain more semantic and global information, we convolve and upsample them to obtain our low-frequency images. 
We use the residual branch ($C-res$) to explicitly combine the high-frequency low-frequency features. 
Lastly, we add these feature images pixel by pixel to get a reconstructed image.

\begin{figure}[!ht]
	\centering
	\includegraphics[width=\linewidth]{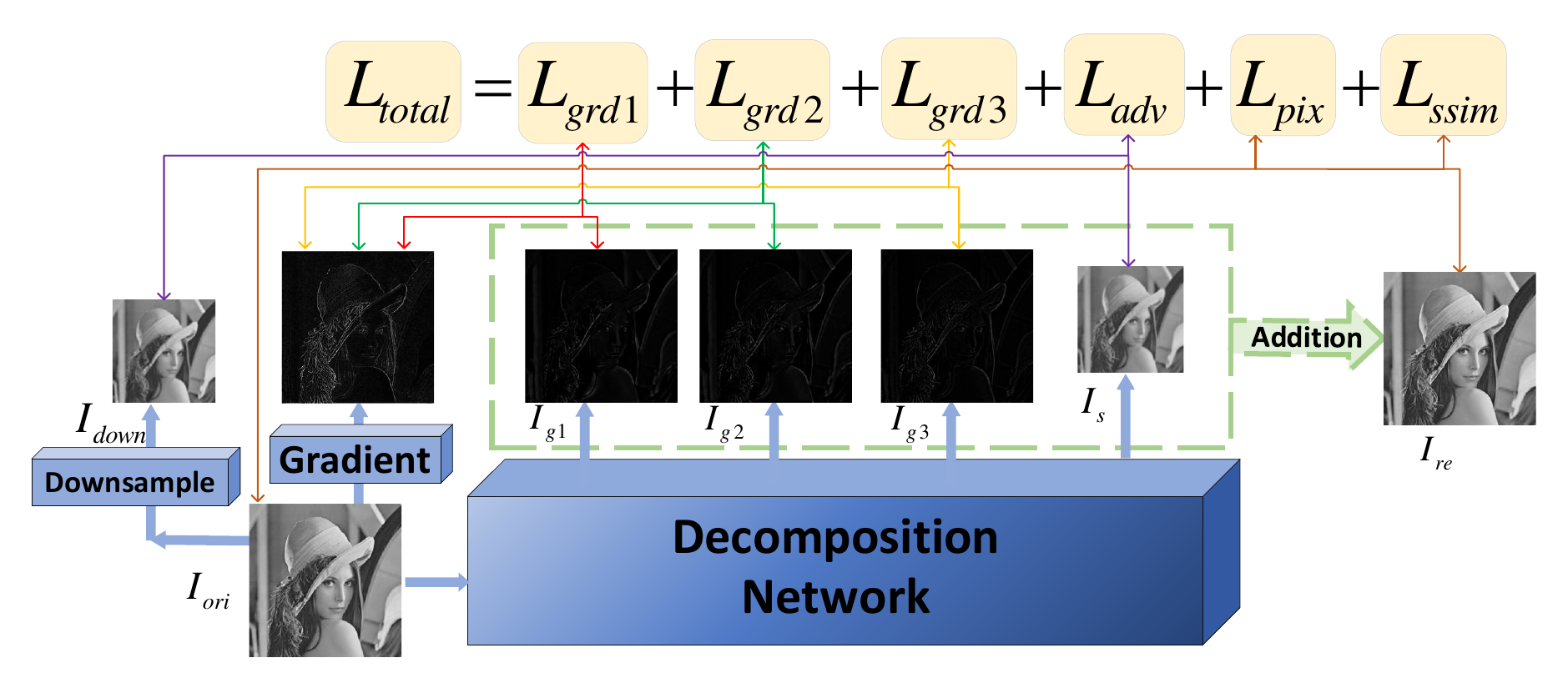}
	\caption{The components of the loss function.}
	\label{fig:loss}
	\vspace{-6pt}
\end{figure}

\subsection{Loss Function}
\label{subsection:Loss}

\begin{figure}[!t]
	\centering
	\resizebox{7cm}{!}{
		\includegraphics[width=\linewidth]{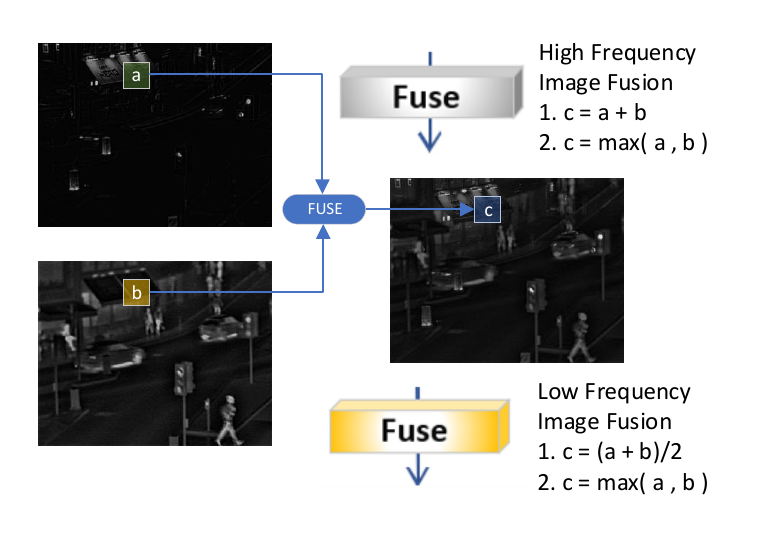}
	}
	\caption{Decomposition and fusion of infrared and visible images.}
	\label{fig:fuse}
	\vspace{-6pt}
\end{figure}
In the training phase, the loss function (${Loss}_{total}$) of our network consists of three components. 
They are the gradient loss (${L}_{detail}$) of the high-frequency image, the distribution loss (${L}_{semantic}$) of the low-frequency image, and the content reconstruction loss (${L}_{reconstruction}$) of the reconstructed image. 
The formula of the loss function is defined as follows:
\begin{equation}
\label{equ:total}
\begin{aligned}
{{Loss}_{total}}={{L}_{detail}}+ \alpha {{L}_{semantic}} + \beta {{L}_{reconstruction}}
\end{aligned}
\end{equation}
$\alpha$ and $\beta$ are hyper-parameters balancing the three losses.

As shown in Fig. \ref{fig:loss}, ${L}_{detail}$ calculates the mean square error loss between the high-frequency feature map ($I_{g1}$, $I_{g2}$, $I_{g3}$) and the gradient the input image and  accumulates them. 
The loss, $L_{detail}$, is formally defined as 

\begin{equation}
\begin{aligned}
&{{L}_{detail}}={L_{grd\text{-}1}}+{L_{grd\text{-}2}}+{L_{grd\text{-}3}}\\
{{L}_{grd\text{-}i}}&=MSE(Gradient({{I}_{ori}}),{{I}_{g\text{-}i}}),i\in \{{1,2,3}\}\\
&MSE(X,Y)=\frac{1}{N}\sum_{n=1}^{N}(X_n - Y_n )^2 
\end{aligned}
\end{equation}
where $I_{ori}$ is the input source image and $I_{g\text{-}i}$ is the $ith$ high-frequency image. 
The $MSE(X,Y)$ is the mean square error between $X$ and $Y$. 
The gradient image of the original image is obtained by using the Laplacian gradient operator $Gradient()$. 
The Laplacian operator performs a mathematical convolution operation according to Equ.\ref{equ:mc}.


In Equ.\ref{equ:total}, ${L}_{semantic}$ is a data distribution loss. 
It combines the loss of the high-frequency image, and the loss of the reconstructed image. 
We define reconstruction loss and detail loss as strong loss and semantic loss as weak loss.
The reconstruction loss constrains the final generated image, and the detail loss constrains the high frequency features of the image. 
In addition, the final feature results consists of high-frequency features and low-frequency features.
The mean square error loss of the reconstruct loss and detail loss is a strong constraint loss at the pixel level. 
If the low-frequency semantic loss is also a strong loss, then it is the sum of three strong losses. Each loss has a pixel-level strong constraint on the image data, which will lead to conflicts and contradictions.
Therefore, we use the mean square error function to constrain the loss of high frequency information and reconstruction. Then, the downsampled image is used as the beacon of low frequency image data distribution, and its data distribution is guided by the adversarial loss function .

\begin{equation}
\begin{aligned}
{{L}_{semantic}}=L_{adv}(I_{s},I_{down})
\end{aligned}
\end{equation}
where $I_{s}$ is the low-frequency semantic image generated by the network, $I_{down}$ is the low-frequency blurred image obtained by downsampling the source image twice, and $L_{adv}$ is the adversarial loss.

\begin{equation}
\label{equ:adv}
\begin{aligned}
L_{adv}(I_{s},I_{down})=L_G=\frac{1}{N}\sum_{n=1}^{N}\left(D\left(G\left(I_s^n\right)\right)-1\right)^2\\
L_D=\frac{1}{N}\sum_{n=1}^{N}{\left(D\left(I_{down}^n\right)-1\right)^2}+\frac{1}{N}\sum_{n=1}^{N}{\left(D\left(I_{s}^n\right)-0\right)^2}
\end{aligned}
\end{equation}
where $n\in\mathbb{N}_N$,  and $N$ represents the number of images. 
Here we use the loss function defined in LSGAN \cite{mao2017least}.

In Equ. \ref{equ:total}, $L_{reconstruction} $ is the image content reconstruction loss of the reconstructed image. 
It consists of two elements. 
One is the pixel-level reconstruction loss ${L}_{pix}$, and the other is the structural similarity loss ${L}_{ssim}$ as follows:
\begin{equation}
\begin{aligned}
{{L}_{reconstruction}}={{L}_{pix}}+\gamma {{L}_{ssim}}
\end{aligned}
\label{equ:5}
\end{equation}
where $\gamma$ is a hyper-parameter that balances the two losses.
$L_{pixel}$ and $L_{ssim}$ are calculated as follows:
\begin{equation}
\begin{aligned}
&{{L}_{pixel}}=MSE({{I}_{ori}},{{I}_{re}})\\
&{{L}_{ssim}}=1-SSIM({{I}_{ori}},{{I}_{re}}) \\
SSIM\left(x,y\right)&=\ \frac{(2\mu_x\mu_y+c_1)(2\sigma_{xy}+c_2)}{(\mu_x^2+\mu_y^2+c_1)(\sigma_x^2+\sigma_y^2+c_2)}
\end{aligned}
\end{equation}

As shown in the Fig. \ref{fig:loss}, the total loss function $L_{total}$ is given as follows:
\begin{equation}
\label{equ:ts}
\begin{aligned}
{{Loss}_{total}}={{L}_{detail}}+ \alpha {{L}_{semantic}} + \beta {{L}_{reconstruction}}\\
={L_{grd\text{-}1}}+{L_{grd\text{-}2}}+{L_{grd\text{-}3}}+\lambda_1 L_{adv}+\lambda_2{{L}_{pix}}+\lambda_3 {{L}_{ssim}}
\end{aligned}
\end{equation}
$\lambda_1,\lambda_2,\lambda_3$ are the hyper-parameters used for balancing the three losses.

\subsection{Image Fusion}
\label{subsection:Network}
\begin{figure}[!ht]
	\centering
	\includegraphics[width=\linewidth]{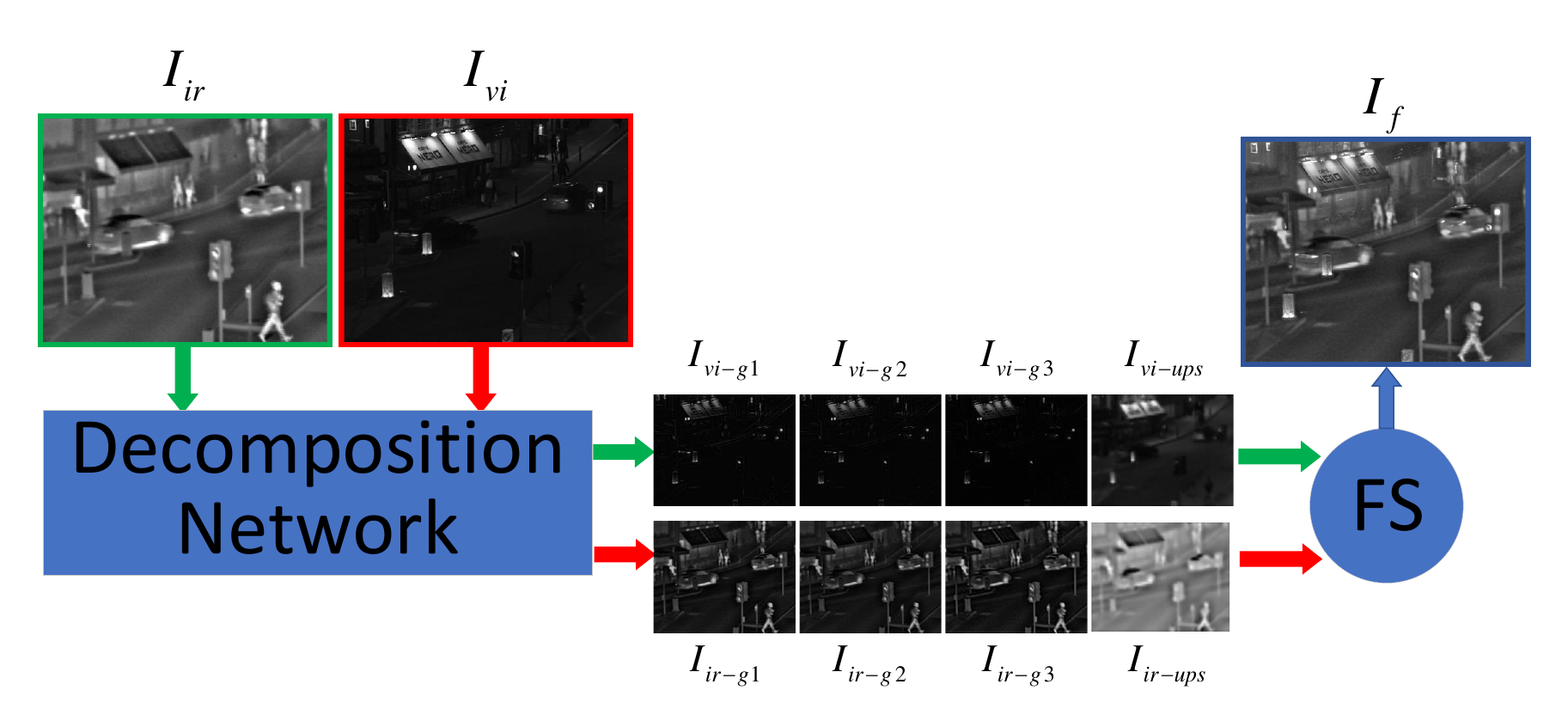}
	\caption{The framework of the proposed method. "Decomposition Network" decomposes the image and "FS" signifies fusion.}
	\label{fig:fusionnetwork}
\end{figure}

Once the system is trained, each test image is first decomposed, as shown in Fig. \ref{fig:fusionnetwork}. 
The fusion strategy ("FS" in Fig.\ref{fig:fusionnetwork})  fuses the corresponding feature images. These are then reconstructed to obtain the final fused image.

In particular, in Fig.\ref{fig:fusionnetwork}, $I_{ir}$ and $I_{vi}$ represent infrared and visible light image respectively. The two images are fed into the decomposition network to obtain two sets of feature images. One group of feature images comes from the visible light images, including three visible light high-frequency images ($I_{vi\text{-}g1}$, $I_{vi\text{-}g2}$, $I_{vi-g3}$) and one visible light low-frequency image ($I_{vi\text{-}ups}$). Another group of feature images comes from infrared images, comprising three infrared high-frequency images ($I_{ir\text{-}g1}$, $I_{ir\text{-}g2}$, $I_{ir\text{-}g3}$) and one infrared low frequency image ($I_{ir\text{-}ups}$). For the corresponding four groups of feature images, our fusion strategy offers a variety of options to obtain the final fused image $I_{f}$ as discussed in the following subsection.

\subsection{Fusion strategy}

We design a fusion strategy to get a fused image. As shown in the Fig.\ref{fig:fuse}, we first use the decomposition network to decompose the visible light image $I_{vi}$ and the infrared image $I_{ir}$ to obtain two sets of high and low frequency feature images. The corresponding high-frequency and low-frequency feature images (such as $I_{vi\text{-}g1}$ and $I_{ir\text{-}g1}$) are fused using different specific  fusion strategies to obtain fused high-frequency feature  images and low-frequency feature  images ($I_{f\text{-}g1}$, $I_{f\text{-}g2}$, $I_{f\text{-}g3}$, $I_{f\text{-}ups}$). Finally, the fusion feature image is added pixel by pixel to obtain the fused image $I_{f}$, which is the same as reconstructing an image in the training phase.

\begin{figure}[!ht]
	\centering
	\includegraphics[width=85mm]{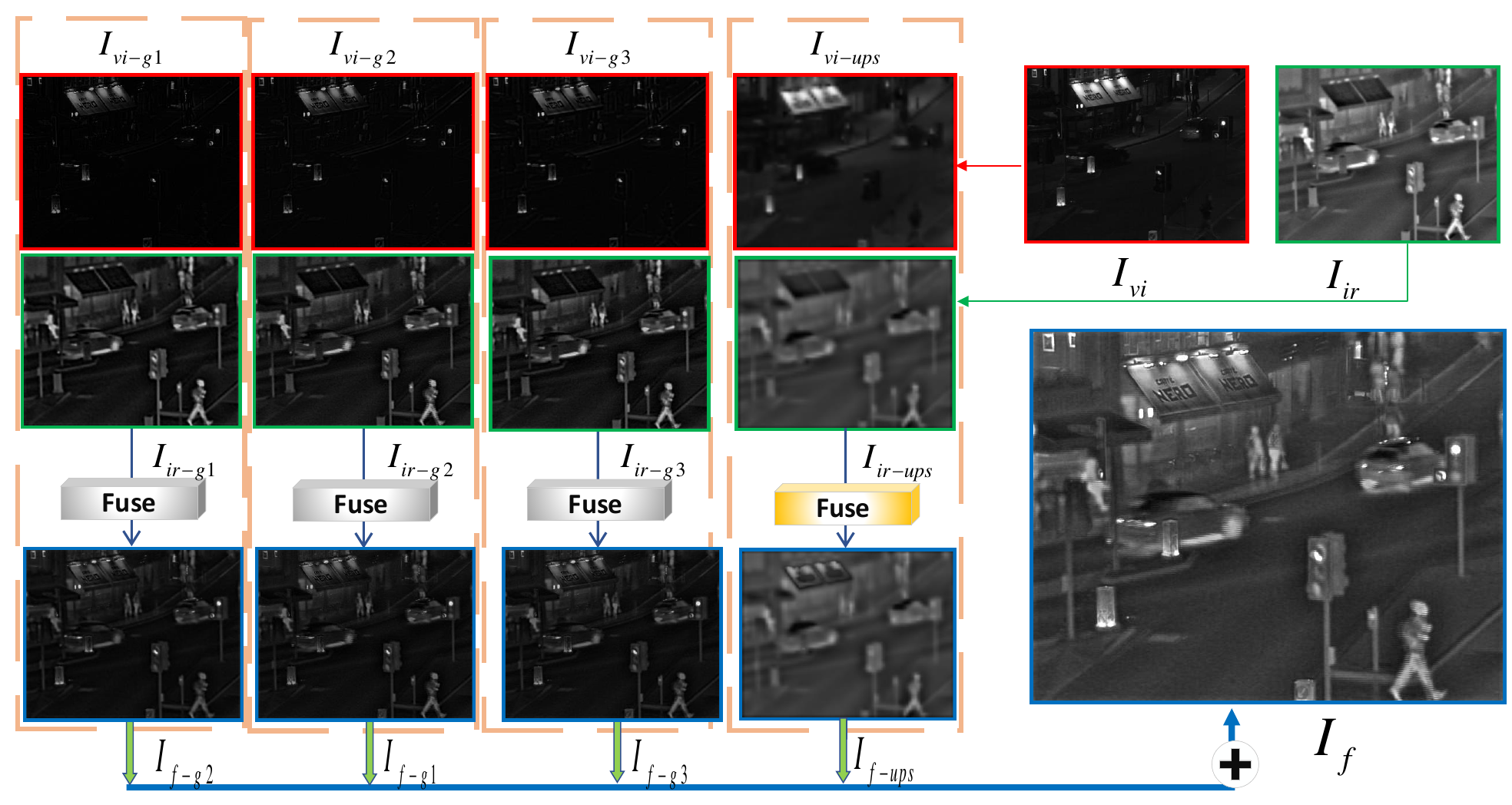}
	\caption{High and low frequency image fusion strategy. Here a and b are any pair of points from feature images, and c is the corresponding fused pixel.}
	\label{fig:strategy}
\end{figure}

We used two fusion strategies for both high-frequency image fusion, and low-frequency image fusion, namely, pixel-wise averaging (avg) and pixel wise max pooling (max) as shown in Fig.\ref{fig:strategy}.
The formulas for high frequency feature fusion $I_{f\text{-}g}$ and low frequency feature fusion $I_{f\text{-}ups}$ are given as follows:
\begin{equation}
\begin{aligned}
& I_{f\text{-}gj}^{n}=max(I_{vi\text{-}gi}^{n},I_{ir\text{-}gi}^{n}) \ , &(Max)\\ 
 or\ & 
I_{f\text{-}gi}^{n}=I_{vi\text{-}gi}^{n}+I_{ir\text{-}gj}^{n} &(Add)\\ \\ 
& I_{f\text{-}ups}^{n}=\frac{(I_{vi\text{-}ups}^{n},I_{ir\text{-}ups}^{n})}{2} \ ,&(Avg)\\ 
 or\ &
 I_{f\text{-}ups}^{n}=max(I_{vi\text{-}ups}^{n},I_{ir\text{-}ups}^{n}) &(Max)\\
  &i\in \{{1,2,3}\} , n\in \mathbb{N}
\end{aligned}
\end{equation}
where $i$ represents the high-frequency image index, and $\mathbb{N}$ denotes the number of pixels in the image. $I_{vi\text{-}gi}^{n}$ and $I_{ir\text{-}gi}^{n}$ represent a pixel in the corresponding three groups of high-frequency images, and $I_{vi\text{-}ups}^{n}$ and $I_{ir\text{-}ups}^{n}$ are the pixels in the low-frequency image. We calculate and fuse the corresponding pixels to produce the fused high frequency image $I_{f\text{-}gi}^{n}$ and low frequency image $I_{f\text{-}ups}^{n}$.
Finally, the three fusion features are added to obtain the final fused image $I_{f}$ as follows:
\begin{equation}
{{I}_{f}}={{I}_{f-g1}}+{{I}_{f-g2}}+{{I}_{f-g3}}+{{I}_{f-ups}}
\end{equation}
\label{subsection:Fusion}

\section{EXPERIMENTS AND ANALYSIS}
\label{section:EXPERIMENTS}

\subsection{Training and Testing Details}

The choice of hyper-parameters is guided by the requirement to maintain the contributions to the final loss values of the same order of magnitude. Accordingly, in formula \ref{equ:ts}, we set $\lambda_1$ = 0.1, $\lambda_2$ = 100, $\lambda_3$ = 10 by cross validation.

Our goal is to train a powerful decomposition network that can decompose images into high-frequency and low-frequency components. In this sense, our input images in the training phase are not limited to multi-modal image data. We can also use MS-COCO\cite{lin2014microsoft} and Imagenet\cite{deng2009imagenet} or other images to achieve this goal.
In our experiment, we use MS-COCO as the training set to design our decomposition network. We select about 80,000 images as input images. These images are converted to gray scale images which are then resized to 256$\times $256.

For the infrared image and visible light image fusion task, we use the TNO dataset \cite{toet2014tno} and the RoadScene dataset \cite{xu2020fusiondn}. 
For the RoadScene dataset, we convert the images to gray scale to keep the the visible light image channels consistent with infrared image. 
For the multi-focus image fusion task, we use the Lytro dataset \cite{Nejati2015Multi}. 
The Lytro image are split according to the RGB channels to obtain three pairs of images. 
The fusion result is merged according to the RGB to obtain a fused image.
For the medical image fusion task, we use the Harvard dataset\cite{Harvard}. 
For the multi-focus image fusion task, we use the dataset in  \cite{Cai2018medataset}. 


We input batchsize of 64 images to the network every iteration. And, we select Adam\cite{kingma2014adam} optimiser with an adaptive learning rate decay method\cite{zeiler2012adadelta} as the learning rate scheduler. We set the initial learning rate to 1e-3, the attenuation factor to 0.5, the stopping criterion  to 5 iterations, and the minimum learning rate threshold to 1e-8. We set the maximum number of epoches to 1000.

In the test phase, because our network is fully convolutional, we input infrared images and visible light images without preprocessing operations.

The experiment is conducted on the two NVIDIA TITAN Xp GPUs and 128GB of CPU memory. We decompose 1000 images with 256$\times$256 resolution one by one and calculate the average calculation time. It takes about 2ms to decompose each image.

\subsection{The role of the adversarial loss}
\begin{figure}[!htbp]
	\centering
	\includegraphics[width=\linewidth]{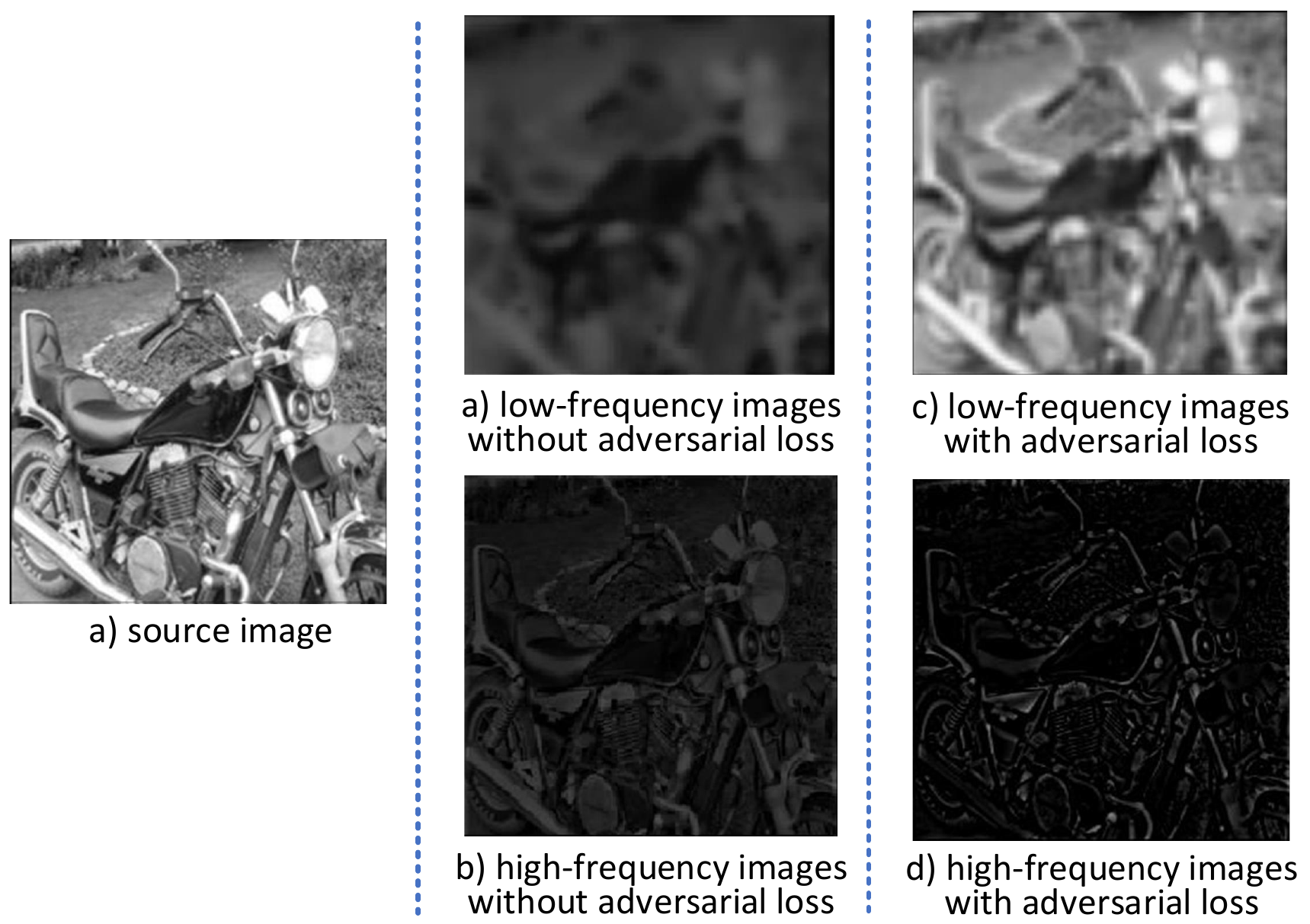}
	\caption{The effect of adversarial loss. a) is the original image, b) and c) are the high-frequency image and low-frequency images without adversarial loss, and d) and e) are the results of applying the adversarial loss.}
	\label{fig:adversarialloss}
\end{figure}
As shown in Fig \ref{fig:adversarialloss}(a), it is difficult for the network to learn to extract the semantic low-frequency content, if we do not impose constraints on the low-frequency image $I_s$. More crucially, low-frequency images tend to lose important semantic information.

As shown in Fig \ref{fig:adversarialloss}(b), without the ${L}_{semantic}$ adversarial loss, the high-frequency images will contain some information that should not appear, such as the brightness, contrast, colors distribution and other semantic information, which are not high-frequency information. 

In order to allow the $semantic$ block to learn the useful low-frequency information, we guide it using weak supervision loss. As in Equ. \ref{equ:adv}, we regard the down-sampled image $I_{down}$ as an approximate solution of the low-frequency image, so that the low-frequency image $I_s$ generated by the network follows the distribution of low-frequency images.

As shown in \ref{fig:adversarialloss}(c and d), the high-frequency images do not learn wrong low-frequency information, and the low-frequency images contain almost all low-frequency semantic information.

\subsection{The details of the decomposed images }
\begin{figure*}[!htbp]
	\centering
	\includegraphics[width=15cm]{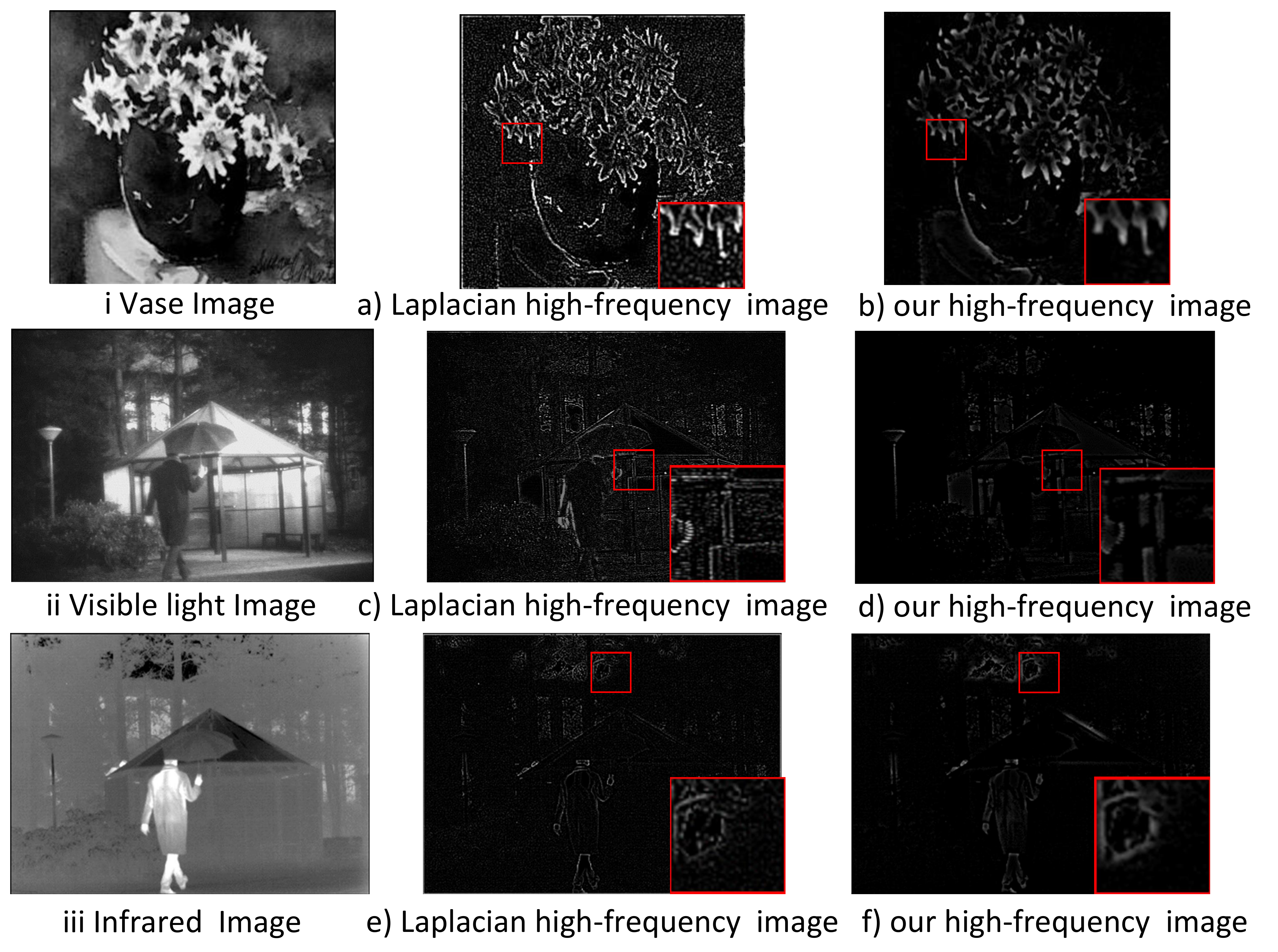}
	\caption{High-frequency images decomposed by the proposed decomposition network and the Laplacian operator.}
	\label{fig:decomposed}
\end{figure*}

Although the loss function of our high-frequency image is calculated with reference to a gradient map produced by the Laplacian operator, there is some discrepancy between them. 
In Fig \ref{fig:decomposed}, we present the high-frequency images decomposed by the proposed decomposition network and the Laplacian operator. It can be seen that the Laplacian gradient images only extract part of the high-frequency information content, and the image has a lot of noise.

The high-frequency image decomposed by our decomposition network not only retains almost all high-frequency information content, but also extracts the object contour and other detailed information. Thus our high-frequency images contain a certain degree of semantic information for recognition.


\subsection{Comparison with State of The Art Methods}
We select ten classic and state of the art fusion methods to compare with our proposed method, including 
Curvelet Transform (CVT)\cite{nencini2007remote}, 
dualtree complex wavelet transform (DTCWT)\cite{lewis2007pixel},  
Laplacian Pyramid (LP)\cite{burt1983the},
Ratio of Low-pass Pyramid (RP)\cite{toet1989image},
DenseFuse\cite{li2018densefuse}, 
the GAN-based fusion network (FusionGAN)\cite{ma2019fusiongan}, 
a general end-to-end fusion network(IFCNN)\cite{zhang2020ifcnn},
ResNetFusion \cite{ma2020infrared}, 
NestFuse\cite{li2020nestfuse}, 
HybridMSD \cite{zhou2016perceptual}, 
PMGI \cite{zhang2020rethinking},
FusionDN\cite{xu2020fusiondn}, 
U2Fusion\cite{xu2020u2fusion},
MDLatLRR\cite{li2020mdlatlrr}, 
Guided Filtering Fusion(GFF)\cite{li2013image},
MEF-GAN\cite{xu2020mef},
Guided Filter focus region detection(GFDF)\cite{QIU2019GFDF},
Convolutional Sparse Representation(CSR)\cite{liu2016image},
Cross Bilateral Filter fusion method(CBF)\cite{kumar2015image},
Discrete Cosine Harmonic Wavelet Transform(DCHWT)\cite{kumar2013multifocus} and
Multi-resolution Singular Value Decomposition (MSVD)\cite{naidu2011image}. 
We use the publicly available codes of these methods and the parameters recommended by their inventors to obtain the fused images.

As there are currently no generally agreed evaluation indicators to measure the quality of the fused image, we will use several objective measures for a comprehensive comparison, as well as a subjective evaluation.

\subsubsection{Subjective and objective evaluation}
In different fields, for different tasks, everyone has his/her own criteria for judging. We focus on the subjective impression of each picture, such as lightness, fidelity, noise, and clarity etc.
We report the results of all approaches and highlight some specific local areas. 

Subjective feelings are good indicators of the fusion performance. However, they cannot be a basis of objective evaluation. We select fifteen objective evaluation indicators from the popular objective indicators in the literature for a comprehensive evaluation. They are: 
Edge Intensity(EI)\cite{xydeas2000objective},
Cross Entropy(CE)\cite{kumar2013multifocus}, 
SF\cite{eskicioglu1995image}, 
Structural SIMilarity (SSIM) \cite{wang2004image},
modified structural similarity (MS-SSIM)\cite{ma2015perceptual},
Entropy (EN)\cite{roberts2008assessment}, 
Sum of Correlation Coefficients (SCD)\cite{aslantas2015new}, 
Fast Mutual Information ($FMI_w$ and $FMI_{dct}$)\cite{haghighat2014fast},
Mutual Information ($MI_{abf}$)\cite{peng2005feature}, 
Standard Deviation of Image (SD)\cite{rao1997fibre},,
Definition (DF)\cite{desheng2004research}, 
Average gradient (AG)\cite{cui2015detail} 
$N_{abf}$ \cite{kumar2013multifocus},
Revised Mutual Information ($Q_{MI}$)\cite{cvejic2006image}, 
$Q_S$\cite{liu2011objective}, 
Nonlinear Correlation Information Entropy ($Q_{NCIE}$)\cite{inbook}, 
correlation coefficient (CC)\cite{han2008study}, 
Phase Congruency Measurement ($Q_P$)\cite{Zhao2006Performance}
$Q_Y$\cite{yang2008novel}, $Q_W$\cite{piella2003new}, $Q_E$\cite{piella2003new},$Q_{CB}$\cite{chen2009new}
and $Q_G$\cite{xydeas2000objective} respectively.

The objective evaluation indicators are cluster into two categories. One group evaluates the fused image by calculating measures such as  edge(EI), the number of mutations in the image(SF), average gradient(AG), entropy (EN), clarity(DF), Cross Entropy(CE),  $N_{abf}$ and contrast of the image(SD), . 
The other group evaluates the fused image by comparison with the source image. Examples include the mutual information (MI, $FMI_w$ and $FMI_{dct}$),Structural SIMilarity (SSIM, MS-SSIM) and a more complex evaluation method ($Q_G$, $Q_Y$, $Q_E$, $Q_W$, $Q_S$ and $Q_{CB}$).

SCD and CC calculates the correlation coefficients between images. 
SSIM and $Q_S$ calculate the similarity between images.
$FMI_{pixel}$ calculate the mutual information between features. 
$N_{abf}$ represents the ratio of noise added to the final image. 
EN measure the amount of information. 
VIFF is used to measure the loss of image information to the distortion process. 
CrossEntropy and MI measure the degree of information correlation between images.

Among them, the lower the value of the $N^{ab/f}$ and the CE and the higher other values, the better the fusion quality of the approach.

We compare the proposed method with other benchmarking methods. The results of the average values computed over all fused images are shown in Tables. The best value in the quality table is made bold in \textcolor{red}{\textbf{red and bold}}, and the second best value is given in \textbf{\emph{bold and italic}}. 

\subsubsection{Visible and Infrared Image Fusion}
\begin{figure*}[!htbp]
	\centering
	\begin{minipage}[t]{0.48\textwidth}
		\centering
		\includegraphics[width=\textwidth]{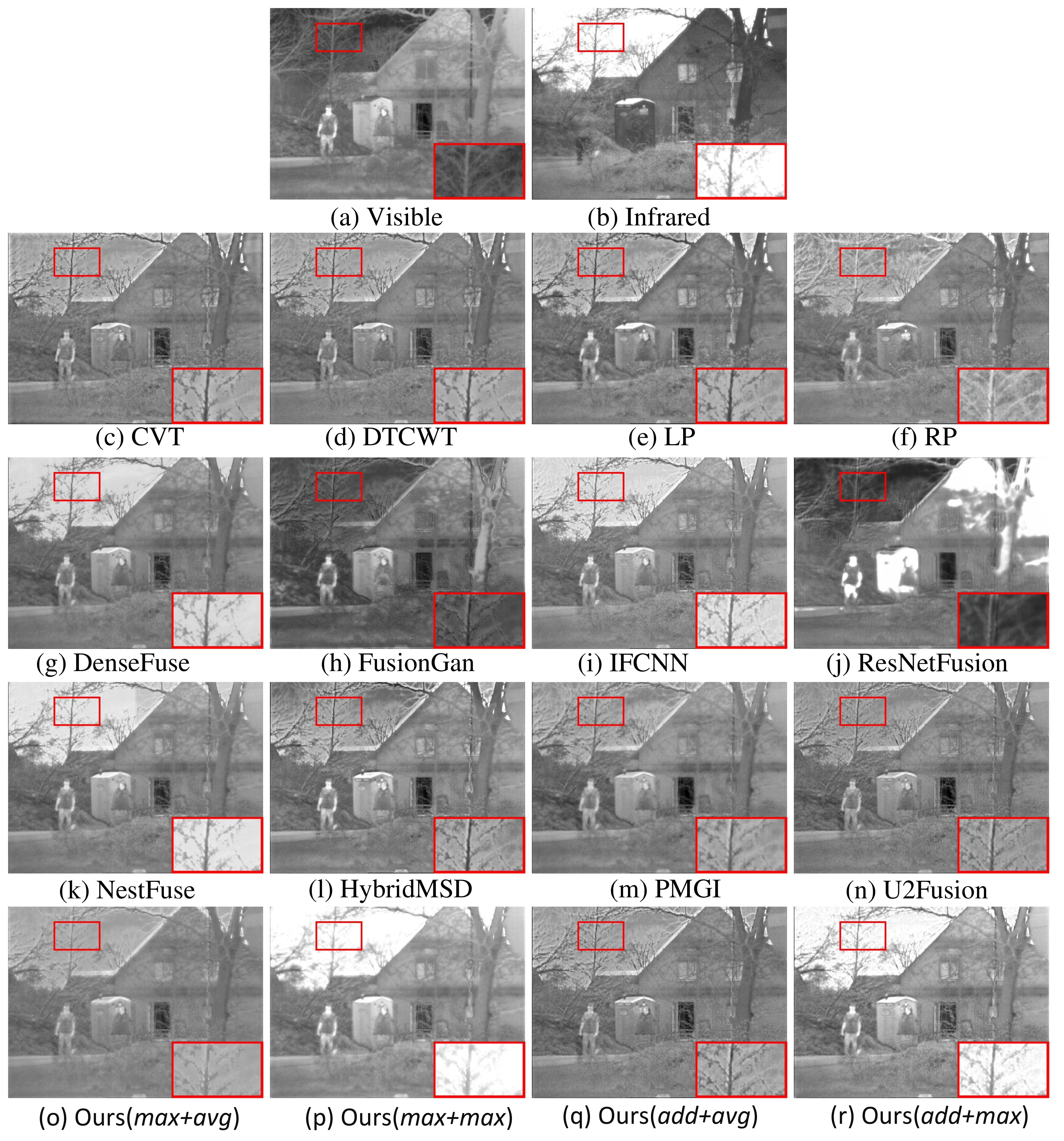}
		\caption{Results in the TNO Dataset.}
		\label{fig:r_tno}
	\end{minipage}
	\begin{minipage}[t]{0.48\textwidth}
		\centering
		\includegraphics[width=\textwidth]{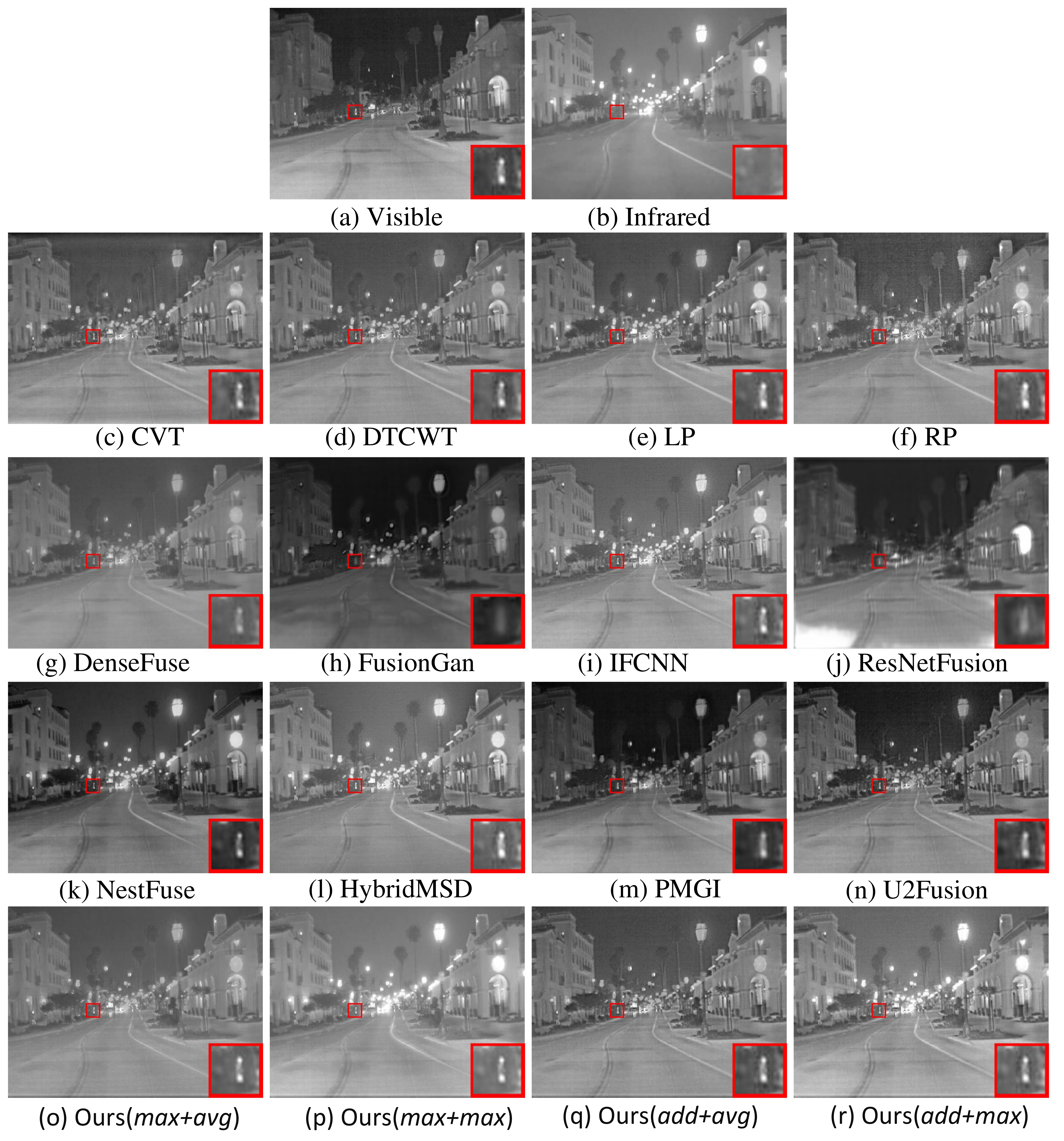}
		\caption{Results in the RoadScene Dataset.}
		\label{fig:r_road}
	\end{minipage}
	\label{fig:rirvi}
\end{figure*}

As shown in Fig. \ref{fig:r_tno}(r) and Fig. \ref{fig:r_road}(q and r), it can be observed that our method has more texture features and is easier to identify small objects.
In TNO dataset, our results can distinguish infrared and visible objects, and reflect more details, such as texture of branches in the sky.
And in the RoadScene dataset, the distant people in our results have a clearer  outline.

As shown in Table \ref{tab:tno}, it can be seen that our results rank in top 2 in multiple indicators. 
Other indicators are also better than most methods. 
It can be demonstrated that our method maintains a effective structural similarity with the source images, preserving a large information correlation with the source images, without introducing noise, artifacts, etc.

Comprehensive analysis from subjective and objective evaluation shows that the $"add+max"$ strategy can achieve better results for infrared and visible light image fusion tasks.

\subsubsection{Multi-exposure Image Fusion}
As shown in Fig. \ref{fig:rexposure}(m and k), it can be observed from the green salient box that our results can well fuse the details of small objects in different exposure environments. Most methods have completely lost the detail information. At the same time, our results look clearer for the exposure and contrast fusion of the whole environment.
As shown in the Table \ref{tab:exposure}, our method can achieve the best results in some indicators, and other indicators are more than half of the methods.

Considering comprehensively, for multi-exposure image fusion task, our method with "add+avg" strategy can get better results .

\subsubsection{Medical Image Fusion}
\begin{figure*}[!htb]
	\centering\resizebox{13cm}{!}{
		\includegraphics[width=\linewidth]{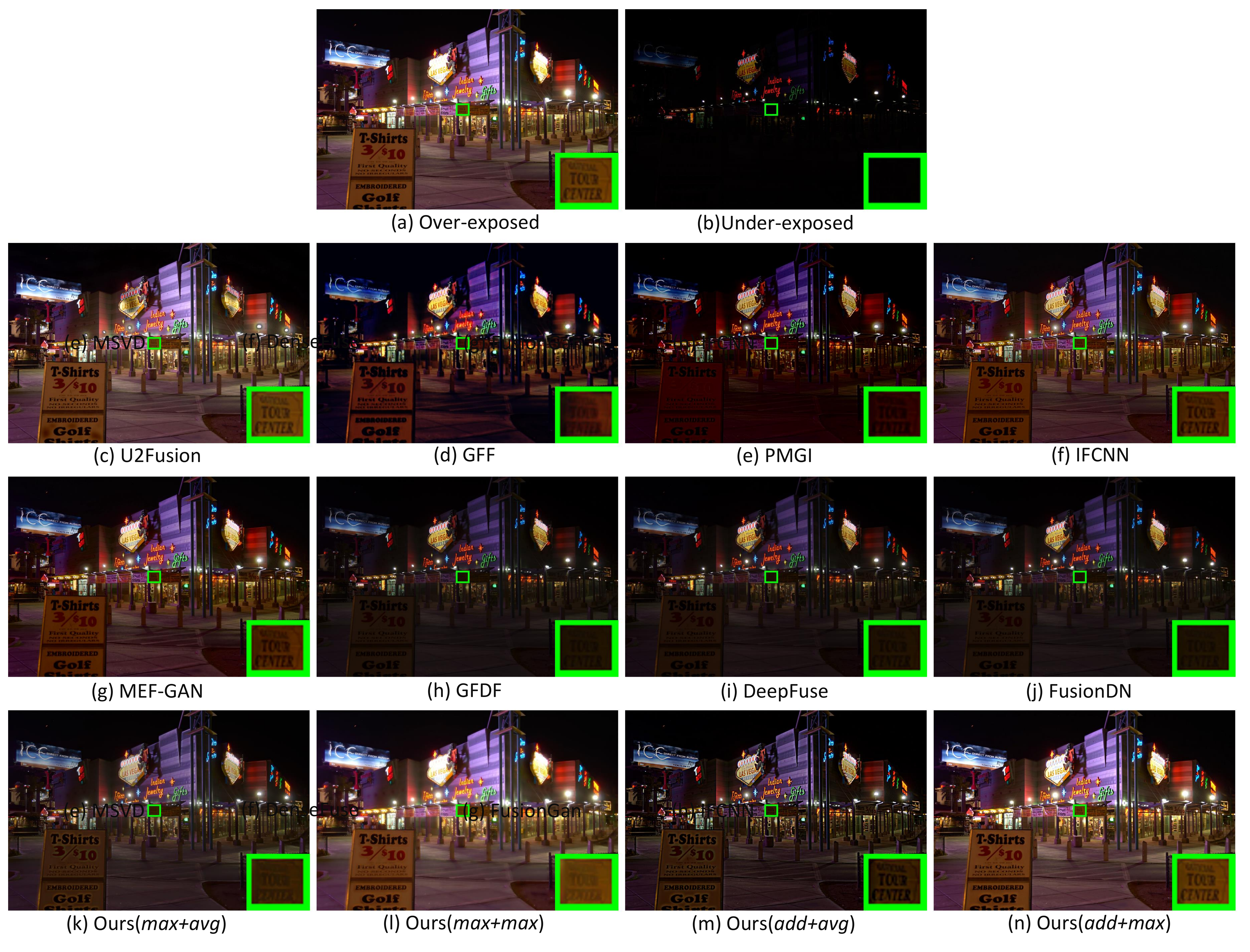}
	}
	\caption{Results in the HDR Dataset.}
	\label{fig:rexposure}
\end{figure*}

As shown in Fig. \ref{fig:rexposure}(k and i), the texture information of brain folds in MRI images is preserved in our medical image results, while other methods lose a lot of texture information. Although the distribution of PET very important, the MRI texture information is more important. The (k)(add+avg) result perfectly fuses the information of both, but the PET information of the (i)(add+max) result is a little distorted.

It can be seen from Table \ref{tab:med} that the "add+avg" and "add+max" strategies have obtained the top two of most indicators. But considering the influence of subjective evaluation, on the whole, using "add+avg" strategy can better accomplish the medical image fusion task.

\subsubsection{Multi-focus Image Fusion}
\begin{figure*}[!htb]
	\centering\resizebox{13cm}{!}{
		\includegraphics[width=\linewidth]{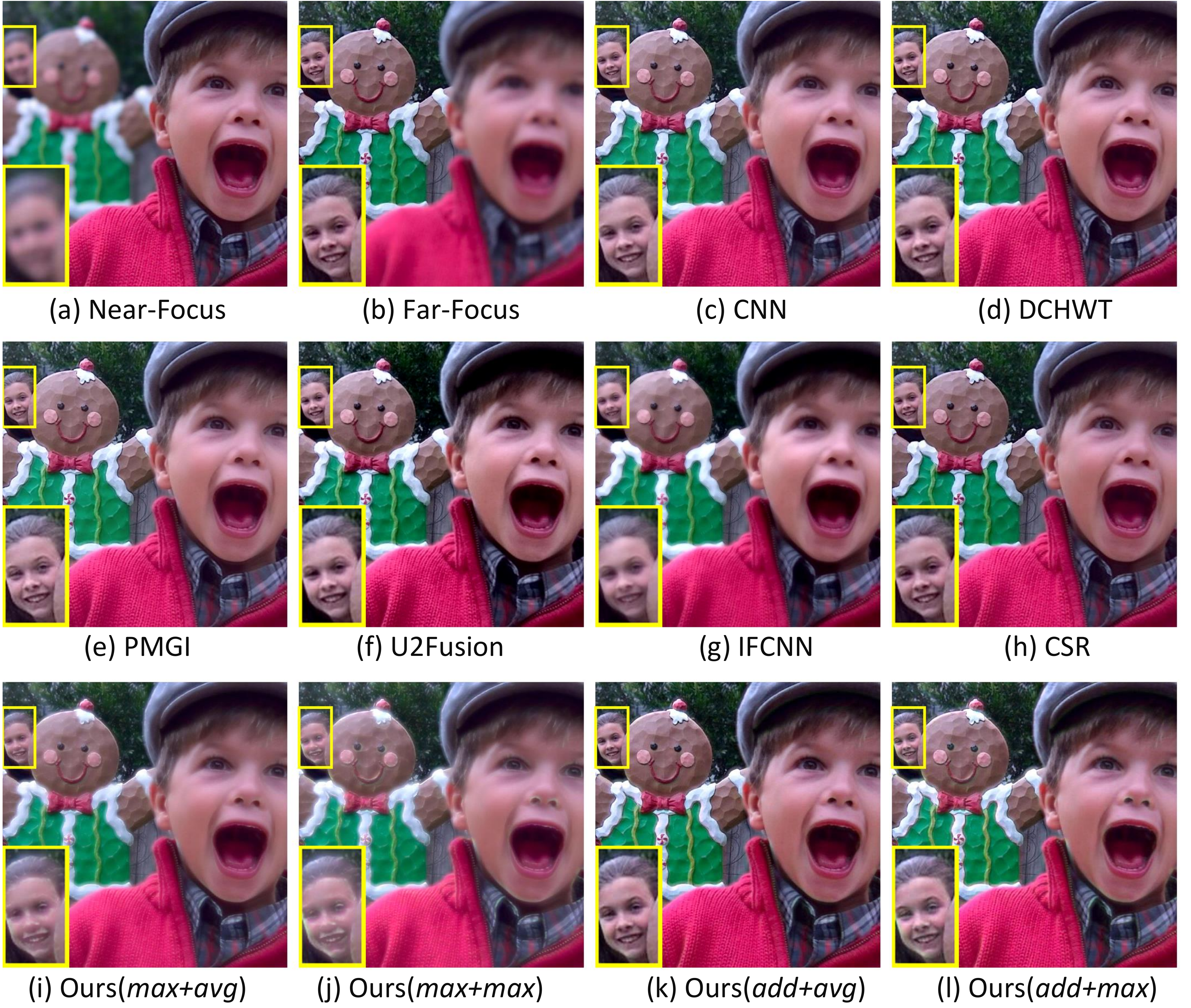}
	}
	\caption{Results in the LYTRO Dataset.}
	\label{fig:rlytro}
\end{figure*}

As shown in Fig. \ref{fig:rlytro}(k), using add+avg strategy, our result image can capture images with different focal lengths very clearly.
As shown in Table \ref{tab:lytro}, it can be seen that our results rank in top 2 in multiple indicators. 
Other indicators are also better than most methods. 

Especially, the result with "add+avg" strategy is the best in several indicators.
Obviously, for the multi-focus image fusion task, our method using "add+avg" strategy can get better results compared with other methods.

\subsubsection{Image Fusion}
We do experiments on four different image fusion tasks on five datasets. It is concluded that our method with "add+avg" and "add+max" strategy can accomplish most of the fusion tasks and get better results.
This is also a matter of course, because the high frequency information of different source images should be different.
For example, details of infrared images and visible images describe different texture features of different objects. More notably, when applied to multi-focus image tasks, the details are not distributed in same place. Therefore, it is feasible and effective to use addition operation for high frequency information.

\begin{figure*}[!htb]
	\centering\resizebox{13cm}{!}{
		\includegraphics[width=\linewidth]{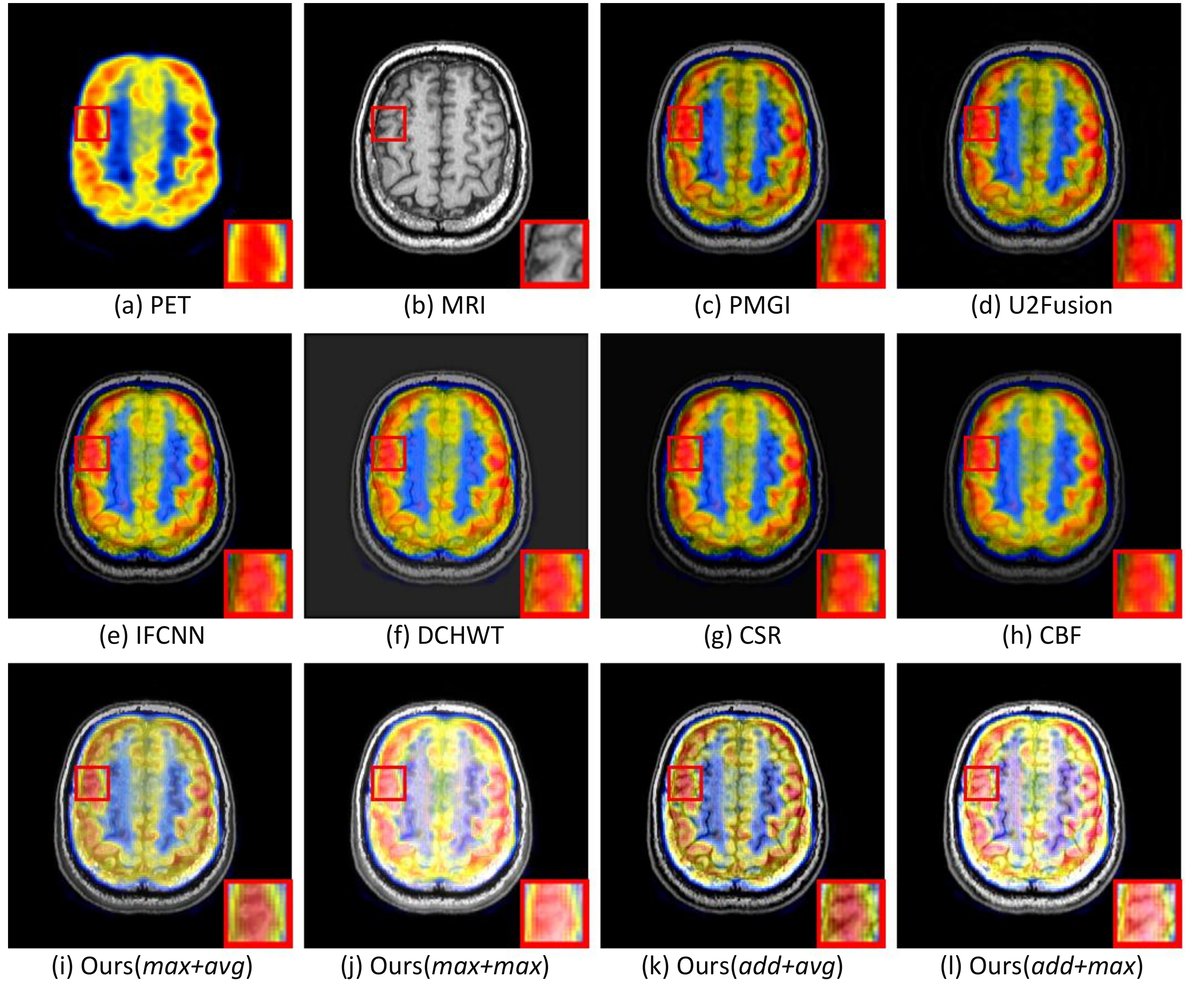}
	}
	\caption{Results in the HARVARD MEDICAL Dataset.}
	\label{fig:rmedical}
\end{figure*}

\begin{table*}[!htb]
	\centering
	\caption{\label{tab:tno}OBJECTIVE EVALUATION OF CLASSICAL AND LATEST FUSION ALGORITHMS ON THE TNO DATASET  }
	\resizebox{\linewidth}{!}{
	\begin{tabular}{@{}cccccccccccccccc@{}}
	\toprule
	&Methods  & EI & CE & SF & SCD & $FMI_{dct}$ & SSIM & $MS_{SSIM}$ & $N_{abf}$ & SD & DF & $Q_{MI}$ & $Q_{NCIE}$ & AG & $MI_{abf}$ \\ \midrule
	\multicolumn{2}{c}{CVT} & 42.9631 & 1.5894 & 11.1129 & 1.5812 & 0.3945 & 0.7025 & 0.8963 & 0.0275 & 27.4613 & 5.4530 & 0.2267 & 0.8033 & 4.2802 & 1.4707 \\
	\multicolumn{2}{c}{DTCWT} & 42.4889 & 1.6235 & 11.1296 & 1.5829 & 0.3936 & 0.7057 & 0.9053 & 0.0232 & 27.3099 & 5.4229 & 0.2385 & 0.8034 & 4.2370 & 1.5428 \\
	\multicolumn{2}{c}{LP} & 44.7055 & 1.4291 & 11.5391 & 1.5920 & 0.3558 & 0.7037 & {\color[HTML]{FF0000} \textit{\textbf{0.9404}}} & 0.0237 & 30.5623 & 5.7422 & 0.2420 & 0.8035 & 4.4690 & 1.5845 \\
	\multicolumn{2}{c}{RP} & 44.9054 & 1.3420 & {\color[HTML]{FF0000} \textit{\textbf{12.7249}}} & 1.5769 & 0.2631 & 0.6705 & 0.8404 & 0.0583 & 28.8385 & \textit{\textbf{6.1799}} & 0.2227 & 0.8032 & 4.6013 & 1.4390 \\
	\multicolumn{2}{c}{DenseFuse} & 36.4838 & 1.4015 & 9.3238 & 1.5329 & 0.3897 & 0.7108 & 0.8692 & 0.0352 & 38.0412 & 4.6176 & 0.4487 & 0.8077 & 3.6299 & 3.0035 \\
	\multicolumn{2}{c}{FusionGan} & 32.5997 & 1.9353 & 8.0476 & 0.6876 & \textit{\textbf{0.4142}} & 0.6235 & 0.6135 & 0.0352 & 29.1495 & 4.2727 & 0.2766 & 0.8039 & 3.2803 & 1.7975 \\
	\multicolumn{2}{c}{IFCNN} & 44.9725 & 1.4413 & 11.8590 & 1.6126 & 0.3739 & 0.7168 & 0.9129 & 0.0346 & 33.0086 & 5.9808 & 0.3419 & 0.8051 & 4.5521 & 2.2513 \\
	\multicolumn{2}{c}{ResNetFusion} & 26.2360 & 1.6188 & 5.9182 & 0.1937 & 0.1003 & 0.4560 & 0.3030 & 0.0550 & {\color[HTML]{FF0000} \textit{\textbf{46.9280}}} & 2.5853 & 0.1211 & 0.8027 & 2.4010 & 0.8014 \\
	\multicolumn{2}{c}{NestFuse} & 37.5627 & 1.4463 & 9.5383 & 1.5742 & 0.3702 & 0.7057 & 0.8816 & 0.0428 & 37.3780 & 4.7890 & \textit{\textbf{0.4672}} & \textit{\textbf{0.8077}} & 3.7475 & \textit{\textbf{3.1049}} \\
	\multicolumn{2}{c}{HybridMSD} & 46.3200 & \textit{\textbf{1.3157}} & 12.3459 & 1.5773 & 0.3207 & 0.7094 & \textit{\textbf{0.9276}} & 0.0435 & 31.2742 & 6.0954 & 0.2567 & 0.8039 & 4.6878 & 1.6987 \\
	\multicolumn{2}{c}{PMGI} & 37.2133 & 1.5656 & 8.7194 & 1.5738 & 0.3949 & 0.6976 & 0.8684 & 0.0340 & 33.0167 & 4.4328 & 0.3021 & 0.8046 & 3.6223 & 2.0237 \\
	\multicolumn{2}{c}{U2Fusion} & \textit{\textbf{48.4915}} & 1.3255 & 11.0368 & 1.5946 & 0.3381 & 0.6758 & 0.9147 & 0.0800 & 31.3794 & 5.8343 & 0.2490 & 0.8035 & \textit{\textbf{4.7392}} & 1.6460 \\ \cmidrule(lr){2-2}
	& max+avg & 29.8401 & 1.7925 & 7.9033 & 1.5954 & 0.2618 & {\color[HTML]{FF0000} \textit{\textbf{0.7253}}} & 0.8554 & {\color[HTML]{FF0000} \textit{\textbf{0.0083}}} & 25.3466 & 3.9337 & 0.2661 & 0.8037 & 3.0034 & 1.7100 \\
	& max+max & 33.3120 & 1.3613 & 8.3395 & 1.6093 & 0.2576 & 0.6911 & 0.8347 & \textit{\textbf{0.0183}} & 38.5507 & 4.1666 & {\color[HTML]{FF0000} \textit{\textbf{0.4812}}} & {\color[HTML]{FF0000} \textit{\textbf{0.8079}}} & 3.2861 & {\color[HTML]{FF0000} \textit{\textbf{3.1549}}} \\
	& add+avg & 46.0796 & 1.4945 & 12.0109 & \textit{\textbf{1.6364}} & {\color[HTML]{FF0000} \textit{\textbf{0.4171}}} & \textit{\textbf{0.7193}} & 0.9196 & 0.0457 & 27.5005 & 6.1199 & 0.2396 & 0.8034 & 4.6589 & 1.5612 \\
	\multirow{-4}{*}{Ours} & add+max & {\color[HTML]{FF0000} \textit{\textbf{48.6587}}} & {\color[HTML]{FF0000} \textit{\textbf{1.2340}}} & \textit{\textbf{12.3736}} & {\color[HTML]{FF0000} \textit{\textbf{1.6547}}} & 0.4075 & 0.6962 & 0.8978 & 0.0565 & \textit{\textbf{39.9552}} & {\color[HTML]{FF0000} \textit{\textbf{6.2775}}} & 0.3618 & 0.8057 & {\color[HTML]{FF0000} \textit{\textbf{4.8675}}} & 2.4344 \\ \bottomrule
	\end{tabular}
}
\end{table*}

\begin{table*}[!htb]
	\centering
	\caption{\label{tab:road}OBJECTIVE EVALUATION OF CLASSICAL AND LATEST FUSION ALGORITHMS ON THE  ROADSCENE DATASET  }
	\resizebox{\linewidth}{!}{
\begin{tabular}{cccccccccccccccc}
	\toprule
	 &Methods   & EI & CE & SF & SCD & $FMI_{dct}$ & SSIM & $MS_{SSIM}$ & $N_{abf}$ & SD & DF & $Q_{MI}$ & $Q_{NCIE}$ & AG & $MI_{abf}$ \\ \midrule
	\multicolumn{2}{c}{CVT} & 59.7642 & 1.1498 & 14.7379 & 1.3418 & 0.3631 & 0.6641 & 0.8721 & 0.0318 & 36.0884 & 6.9618 & 0.2994 & 0.8056 & 5.7442 & 2.1473 \\
	\multicolumn{2}{c}{DTCWT} & 57.3431 & 1.2475 & 14.7318 & 1.3329 & 0.2383 & 0.6567 & 0.8725 & 0.0412 & 34.7264 & 6.7810 & 0.3280 & 0.8060 & 5.5228 & 2.3358 \\
	\multicolumn{2}{c}{LP} & 59.5437 & 1.1479 & 15.3634 & 1.3617 & 0.3092 & 0.6875 & {\color[HTML]{FF0000} \textit{\textbf{0.9110}}} & 0.0250 & 37.3478 & 7.1062 & 0.3432 & 0.8064 & 5.7627 & 2.4641 \\
	\multicolumn{2}{c}{RP} & 50.8057 & 1.1250 & {\color[HTML]{FF0000} \textit{\textbf{19.1529}}} & 1.2829 & 0.2624 & 0.6341 & 0.8297 & 0.0773 & 38.4519 & {\color[HTML]{FF0000} \textit{\textbf{8.8084}}} & 0.3024 & 0.8057 & 6.1410 & 2.1717 \\
	\multicolumn{2}{c}{DenseFuse} & 34.0135 & 1.0665 & 8.5541 & 1.3491 & \textit{\textbf{0.3801}} & 0.6928 & 0.8529 & 0.0125 & 44.0963 & 5.6010 & 0.4309 & 0.8079 & 3.2740 & 3.0226 \\
	\multicolumn{2}{c}{FusionGan} & 35.4048 & 1.8991 & 8.6400 & 0.8671 & \textit{\textbf{0.3609}} & 0.6142 & 0.7352 & 0.0168 & 42.3040 & 3.9243 & 0.4067 & 0.8077 & 3.3469 & 2.9564 \\
	\multicolumn{2}{c}{IFCNN} & 57.6653 & 1.1486 & 15.0677 & 1.3801 & 0.3456 & 0.6746 & 0.8798 & 0.0315 & 35.8183 & 7.0401 & 0.4212 & 0.8076 & 5.6242 & 2.9980 \\
	\multicolumn{2}{c}{ResNetFusion} & 39.4317 & 1.1539 & 8.4967 & 0.2179 & 0.1034 & 0.3599 & 0.2363 & 0.0550 & {\color[HTML]{FF0000} \textit{\textbf{66.8924}}} & 3.8041 & 0.2316 & 0.8046 & 3.6011 & 1.6695 \\
	\multicolumn{2}{c}{NestFuse} & 53.9286 & 1.3026 & 14.2820 & 1.2597 & 0.3484 & 0.6679 & 0.8272 & 0.0432 & 48.9920 & 6.2840 & \textit{\textbf{0.5141}} & \textit{\textbf{0.8098}} & 5.1834 & {\color[HTML]{FF0000} \textit{\textbf{3.7602}}} \\
	\multicolumn{2}{c}{HybridMSD} & 62.2138 & \textit{\textbf{1.1032}} & 16.4475 & 1.2642 & 0.2887 & {\color[HTML]{FF0000} \textit{\textbf{0.6961}}} & \textit{\textbf{0.9127}} & 0.0460 & 37.1333 & 7.5600 & 0.3429 & 0.8061 & 6.0725 & 2.4377 \\
	\multicolumn{2}{c}{PMGI} & 47.2067 & 1.6395 & 10.9368 & 1.0989 & 0.3644 & 0.6640 & 0.8926 & 0.0146 & \textit{\textbf{49.3262}} & 5.1288 & 0.4657 & {\color[HTML]{FF0000} \textit{\textbf{0.8100}}} & 4.4426 & 3.4343 \\
	\multicolumn{2}{c}{U2Fusion} & {\color[HTML]{FF0000} \textit{\textbf{66.2529}}} & 1.4806 & 15.8242 & 1.3551 & 0.3199 & 0.6813 & \textit{\textbf{0.9150}} & 0.0671 & 42.9368 & 7.5930 & 0.3929 & 0.8075 & {\color[HTML]{FF0000} \textit{\textbf{6.3133}}} & 2.8624 \\ \cline{2-2}
	& max+avg & 39.5260 & 1.2660 & 10.7887 & 1.3488 & 0.2312 & {\color[HTML]{FF0000} \textit{\textbf{0.6938}}} & 0.8236 & {\color[HTML]{FF0000} \textit{\textbf{0.0095}}} & 31.8339 & 4.8193 & 0.3675 & 0.8065 & 3.8632 & 2.5851 \\
	& max+max & 39.6608 & {\color[HTML]{FF0000} \textit{\textbf{1.0061}}} & 10.6186 & 1.3311 & 0.2217 & 0.6741 & 0.7550 & \textit{\textbf{0.0123}} & 39.6899 & 4.7187 & {\color[HTML]{FF0000} \textit{\textbf{0.5343}}} & {\color[HTML]{FF0000} \textit{\textbf{0.8098}}} & 3.8444 & {\color[HTML]{FF0000} \textit{\textbf{3.7182}}} \\
	& add+avg & \textit{\textbf{64.6144}} & 1.3276 & \textit{\textbf{16.5089}} & {\color[HTML]{FF0000} \textit{\textbf{1.4158}}} & {\color[HTML]{FF0000} \textit{\textbf{0.3879}}} & \textit{\textbf{0.6892}} & {\color[HTML]{FF0000} \textit{\textbf{0.9166}}} & 0.0605 & 35.9280 & \textit{\textbf{7.7619}} & 0.3453 & 0.8063 & \textit{\textbf{6.2778}} & 2.4657 \\
	\multirow{-4}{*}{Ours} & add+max & {\color[HTML]{FF0000} \textit{\textbf{64.1570}}} & {\color[HTML]{FF0000} \textit{\textbf{1.0596}}} & \textit{\textbf{16.2412}} & {\color[HTML]{FF0000} \textit{\textbf{1.4129}}} & 0.3676 & 0.6700 & 0.8446 & 0.0653 & \textit{\textbf{42.3352}} & {\color[HTML]{FF0000} \textit{\textbf{7.6144}}} & 0.3994 & 0.8070 & {\color[HTML]{FF0000} \textit{\textbf{6.2102}}} & 2.8425 \\ \hline
	\end{tabular}
	}
\end{table*}

\begin{table*}[!htb]
	\centering
	\caption{\label{tab:exposure}OBJECTIVE EVALUATION OF CLASSICAL AND LATEST FUSION ALGORITHMS ON THE HDR DATABASE }
	\resizebox{\linewidth}{!}{
	\begin{tabular}{@{}cccccccccccccc@{}}
		\toprule
		\multicolumn{2}{c}{Methods} & EI & CE & SF & $FMI_{w}$ & $FMI_{dct}$ & SSIM & $MS_{SSIM}$ & $N_{abf}$ & DF & $Q_{Y}$ & AG & CC \\ \midrule
		\multicolumn{2}{c}{U2Fusion} & \textit{\textbf{76.2068}} & 3.6551 & 23.1641 & 0.4123 & 0.4475 & 0.5474 & 0.9393 & 0.1107 & 9.4715 & 0.5648 & \textit{\textbf{7.4982}} & 1.0131 \\
		\multicolumn{2}{c}{GFF} & 72.9816 & 2.4105 & 23.2912 & 0.4752 & 0.5094 & 0.5614 & 0.7765 & 0.0444 & 9.7177 & 0.7704 & 7.3722 & 0.6247 \\
		\multicolumn{2}{c}{PMGI} & 69.9908 & 3.4868 & 23.7738 & 0.4883 & 0.5370 & 0.5547 & 0.9361 & 0.1076 & 9.7482 & 0.5954 & 7.2044 & 1.0125 \\
		\multicolumn{2}{c}{IFCNN} & 68.9809 & 3.9597 & \textit{\textbf{25.6184}} & 0.4794 & 0.5202 & \textit{\textbf{0.6024}} & \textit{\textbf{0.9532}} & 0.0585 & \textit{\textbf{10.6159}} & 0.7092 & 7.3755 & 0.9683 \\
		\multicolumn{2}{c}{MEF-GAN} & 58.1997 & 2.6383 & 15.9812 & 0.2598 & 0.2559 & 0.5059 & 0.8269 & 0.0847 & 6.4662 & 0.4287 & 5.5098 & 0.9670 \\
		\multicolumn{2}{c}{GFDF} & 73.5531 & \textit{\textbf{1.8508}} & 24.2111 & 0.4774 & 0.4575 & 0.5750 & 0.8502 & 0.0476 & 9.7459 & \textit{\textbf{0.8349}} & 7.4140 & 0.6347 \\
		\multicolumn{2}{c}{DeepFuse} & 58.4911 & 3.4316 & 19.7907 & 0.4901 & \textit{\textbf{0.5433}} & 0.5814 & {\color[HTML]{FF0000} \textit{\textbf{0.9570}}} & 0.0371 & 7.9679 & 0.6248 & 5.9601 & \textit{\textbf{1.0244}} \\
		\multicolumn{2}{l}{FusionDN} & 68.8311 & 3.2806 & 22.0760 & {\color[HTML]{FF0000} \textit{\textbf{0.5558}}} & 0.5340 & 0.5781 & 0.7903 & 0.0190 & 9.2019 & 0.8081 & 6.9623 & 0.6147 \\ \cmidrule(lr){2-2}
		& max+avg & 43.1195 & 2.2750 & 14.8597 & 0.4296 & 0.4361 & {\color[HTML]{FF0000} \textit{\textbf{0.6037}}} & 0.9108 & {\color[HTML]{FF0000} \textit{\textbf{0.0023}}} & 5.7758 & 0.6302 & 4.3684 & 0.9703 \\
		& max+max & 35.5574 & {\color[HTML]{FF0000} \textit{\textbf{1.2851}}} & 12.1623 & 0.4387 & 0.4060 & 0.5900 & 0.8328 & \textit{\textbf{0.0039}} & 4.5982 & {\color[HTML]{FF0000} \textit{\textbf{0.8444}}} & 3.5538 & {\color[HTML]{FF0000} \textit{\textbf{1.0246}}} \\
		& add+avg & {\color[HTML]{FF0000} \textit{\textbf{79.2416}}} & 3.6129 & {\color[HTML]{FF0000} \textit{\textbf{27.2045}}} & \textit{\textbf{0.5275}} & {\color[HTML]{FF0000} \textit{\textbf{0.5442}}} & 0.5796 & 0.9502 & 0.0932 & {\color[HTML]{FF0000} \textit{\textbf{11.1081}}} & 0.6924 & {\color[HTML]{FF0000} \textit{\textbf{8.1807}}} & 0.9782 \\
		\multirow{-4}{*}{Ours} & add+max & 69.2224 & 2.7025 & 24.1946 & 0.4942 & 0.5006 & 0.5940 & 0.9000 & 0.0671 & 9.6762 & 0.7376 & 7.1423 & 0.9382 \\ \bottomrule
	\end{tabular}
	}
\end{table*}

\begin{table*}[!htb]
	\centering
	\caption{\label{tab:med}OBJECTIVE EVALUATION OF CLASSICAL AND LATEST FUSION ALGORITHMS ON THE HARVARD MEDICAL DATASET  }
	\resizebox{\linewidth}{!}{
	\begin{tabular}{@{}cccccccccccccc@{}}
		\toprule
		\multicolumn{2}{c}{Methods} & EI & SF & $Q_{abf}$ & $FMI_{w}$ & $FMI_{dct}$ & SSIM & $MS_{SSIM}$ & $FMI_{pixel}$ & $N_{abf}$ & VIF & SD & DF \\ \midrule
		\multicolumn{2}{c}{PMGI} & 62.0459 & 18.4579 & 0.1927 & 0.1778 & 0.1468 & 0.1396 & 0.5801 & 0.7838 & 0.0249 & 0.2634 & 52.7224 & 7.2918 \\
		\multicolumn{2}{c}{U2Fusion} & 46.2968 & 14.3793 & 0.1546 & 0.2154 & 0.1712 & 0.2005 & 0.5696 & 0.7981 & 0.0191 & 0.3018 & 57.6757 & 5.2656 \\
		\multicolumn{2}{c}{IFCNN} & 61.5885 & 20.2496 & 0.2006 & 0.2149 & 0.1424 & 0.5780 & 0.6090 & 0.7881 & 0.0309 & 0.3351 & 64.2122 & 7.2945 \\
		\multicolumn{2}{c}{DCHWT} & 57.5602 & 18.3965 & 0.1891 & 0.1607 & 0.1486 & 0.5531 & 0.6085 & 0.7898 & 0.0222 & 0.2919 & 56.6969 & 6.8922 \\
		\multicolumn{2}{c}{CSR} & 61.5017 & 19.1411 & 0.1908 & 0.2071 & 0.1386 & 0.5629 & 0.5902 & 0.7826 & 0.0299 & 0.2871 & 55.2012 & 7.1761 \\
		\multicolumn{2}{c}{CBF} & 61.4869 & 19.1834 & 0.1962 & 0.2157 & 0.1472 & 0.5695 & 0.6012 & 0.7909 & 0.0291 & 0.2983 & 56.2595 & 7.0992 \\ \cmidrule(lr){2-2}
		& max+avg & 71.8719 & 24.7649 & 0.5056 & 0.4066 & 0.2948 & 0.7254 & \textit{\textbf{0.9108}} & 0.8382 & {\color[HTML]{FF0000} \textit{\textbf{0.0044}}} & 0.5158 & 77.8777 & 8.7039 \\
		& max+max & 74.5375 & 25.2957 & 0.5022 & 0.3818 & 0.2888 & 0.7086 & 0.8570 & 0.8396 & \textit{\textbf{0.0058}} & 0.5881 & {\color[HTML]{FF0000} \textit{\textbf{95.7545}}} & 8.8149 \\
		& add+avg & \textit{\textbf{95.9522}} & \textit{\textbf{31.1427}} & \textit{\textbf{0.5906}} & {\color[HTML]{FF0000} \textit{\textbf{0.4866}}} & \textit{\textbf{0.4086}} & {\color[HTML]{FF0000} \textit{\textbf{0.7423}}} & {\color[HTML]{FF0000} \textit{\textbf{0.9298}}} & \textit{\textbf{0.8433}} & 0.0279 & \textit{\textbf{0.6635}} & 77.0983 & \textit{\textbf{11.7871}} \\
		\multirow{-4}{*}{Ours} & add+max & {\color[HTML]{FF0000} \textit{\textbf{98.6637}}} & {\color[HTML]{FF0000} \textit{\textbf{32.1464}}} & {\color[HTML]{FF0000} \textit{\textbf{0.6054}}} & \textit{\textbf{0.4833}} & {\color[HTML]{FF0000} \textit{\textbf{0.4147}}} & \textit{\textbf{0.7408}} & 0.8932 & {\color[HTML]{FF0000} \textit{\textbf{0.8505}}} & 0.0240 & {\color[HTML]{FF0000} \textit{\textbf{0.7373}}} & \textit{\textbf{93.8827}} & {\color[HTML]{FF0000} \textit{\textbf{12.0055}}} \\ \midrule
		\multicolumn{2}{c}{Methods} & $Q_{MI}$ & $Q_{S}$ & $Q_{NCIE}$ & AG & MI & QG & CC & VIFF & $Q_{P}$ & $Q_{W}$ & $Q_{E}$ & $Q_{CB}$ \\ \midrule
		\multicolumn{2}{c}{PMGI} & 0.4978 & 0.1678 & 0.8048 & 6.0292 & 2.2892 & 0.4282 & 0.8613 & 0.1660 & 0.0574 & 0.2029 & 0.0130 & 0.2800 \\
		\multicolumn{2}{c}{U2Fusion} & 0.5786 & 0.2102 & 0.8052 & 4.3710 & 2.4827 & 0.4892 & 0.8718 & 0.1840 & 0.0646 & 0.1611 & 0.0045 & 0.2937 \\
		\multicolumn{2}{c}{IFCNN} & 0.5735 & 0.5399 & 0.8051 & 5.9112 & 2.4056 & 0.5445 & 0.8625 & 0.2029 & 0.0602 & 0.1919 & 0.0138 & 0.5737 \\
		\multicolumn{2}{c}{DCHWT} & 0.4911 & 0.5243 & 0.8046 & 5.5238 & 2.2234 & 0.3233 & 0.8634 & 0.1878 & 0.0624 & 0.2025 & 0.0147 & 0.4463 \\
		\multicolumn{2}{c}{CSR} & 0.5144 & 0.5309 & 0.8046 & 5.8969 & 2.2151 & 0.4894 & 0.8529 & 0.1731 & 0.0489 & 0.1957 & 0.0153 & 0.5661 \\
		\multicolumn{2}{c}{CBF} & 0.5471 & 0.5374 & 0.8048 & 5.8732 & 2.2932 & 0.5429 & 0.8574 & 0.1836 & 0.0536 & 0.2032 & 0.0163 & 0.5740 \\ \cmidrule(lr){2-2}
		& max+avg & 0.5879 & 0.8270 & 0.8056 & 6.9650 & 2.6338 & 0.6631 & {\color[HTML]{FF0000} \textit{\textbf{0.9125}}} & 0.4378 & 0.3115 & 0.7888 & 0.5188 & 0.5909 \\
		& max+max & \textit{\textbf{0.6177}} & 0.7958 & {\color[HTML]{FF0000} \textit{\textbf{0.8061}}} & 7.1472 & {\color[HTML]{FF0000} \textit{\textbf{2.8014}}} & 0.6417 & 0.9025 & 0.4319 & 0.2910 & 0.7865 & 0.5394 & 0.5537 \\
		& add+avg & 0.5766 & {\color[HTML]{FF0000} \textit{\textbf{0.8624}}} & 0.8052 & \textit{\textbf{9.3113}} & 2.4738 & {\color[HTML]{FF0000} \textit{\textbf{0.7313}}} & 0.9090 & \textit{\textbf{0.4509}} & \textit{\textbf{0.3367}} & \textit{\textbf{0.8246}} & \textit{\textbf{0.7171}} & {\color[HTML]{FF0000} \textit{\textbf{0.6485}}} \\
		\multirow{-4}{*}{Ours} & add+max & {\color[HTML]{FF0000} \textit{\textbf{0.6210}}} & \textit{\textbf{0.8604}} & \textit{\textbf{0.8060}} & {\color[HTML]{FF0000} \textit{\textbf{9.5239}}} & \textit{\textbf{2.7425}} & \textit{\textbf{0.7311}} & \textit{\textbf{0.9095}} & {\color[HTML]{FF0000} \textit{\textbf{0.4566}}} & {\color[HTML]{FF0000} \textit{\textbf{0.3642}}} & {\color[HTML]{FF0000} \textit{\textbf{0.8674}}} & {\color[HTML]{FF0000} \textit{\textbf{0.7579}}} & \textit{\textbf{0.6308}} \\ \bottomrule
	\end{tabular}
	}
\end{table*}

\begin{table*}[!htb]
	\centering
	\caption{\label{tab:lytro}OBJECTIVE EVALUATION OF CLASSICAL AND LATEST FUSION ALGORITHMS ON THE LYTRO DATASET  }
	\resizebox{\linewidth}{!}{
	\begin{tabular}{ccccccccccccccc}
		\hline
		\multicolumn{2}{c}{Methods} & EI & CE & SF & EN & $FMI_{w}$ & $FMI_{dct}$ & SSIM & $N_{abf}$ & VIF & EN & DF & AG & VIFF \\ \hline
		\multicolumn{2}{c}{CNN} & 74.8426 & 0.2176 & 20.5655 & 7.5498 & 0.4410 & 0.3845 & 0.8303 & 0.0561 & 1.2668 & 7.5498 & 8.6838 & 7.2248 & 1.0196 \\
		\multicolumn{2}{c}{DCHWT} & 68.8342 & 0.2129 & 19.1235 & 7.5385 & 0.4225 & 0.3959 & \textit{\textbf{0.8438}} & 0.0152 & 1.1401 & 7.5385 & 8.0241 & 6.6455 & 0.9646 \\
		\multicolumn{2}{c}{PMGI} & 54.3821 & 0.3815 & 13.6508 & 7.5314 & 0.3751 & 0.3430 & 0.8196 & 0.0122 & 1.1511 & 7.5314 & 5.8043 & 5.0955 & 0.9634 \\
		\multicolumn{2}{c}{U2Fusion} & 75.6982 & 0.3294 & 19.1046 & 7.5275 & 0.3632 & 0.3104 & 0.7795 & 0.0519 & {\color[HTML]{FF0000} \textit{\textbf{1.4957}}} & 7.5275 & 8.1977 & 7.1401 & {\color[HTML]{FF0000} \textit{\textbf{1.1211}}} \\
		\multicolumn{2}{c}{IFCNN} & 73.2405 & 0.2225 & 20.3514 & 7.5432 & 0.4148 & 0.3716 & 0.8367 & 0.0429 & 1.1924 & 7.5432 & 8.5298 & 7.0835 & 0.9852 \\
		\multicolumn{2}{c}{CSR} & 68.1733 & 0.2762 & 18.6452 & 7.5221 & 0.4153 & 0.3446 & 0.7984 & 0.0346 & 1.2672 & 7.5329 & 8.1495 & 6.4281 & 0.9875 \\ \cline{2-2}
		& max+avg & 50.0733 & {\color[HTML]{FF0000} \textit{\textbf{0.0403}}} & 13.1598 & 7.4942 & 0.3513 & 0.2487 & {\color[HTML]{FF0000} \textit{\textbf{0.8535}}} & {\color[HTML]{FF0000} \textit{\textbf{0.0046}}} & 0.8481 & 7.4942 & 5.5959 & 4.7698 & 0.7489 \\
		& max+max & 49.5689 & \textit{\textbf{0.0765}} & 12.8854 & 7.4911 & 0.3451 & 0.2466 & 0.8404 & \textit{\textbf{0.0059}} & 0.7974 & 7.4911 & 5.5495 & 4.7246 & 0.7112 \\
		& add+avg & {\color[HTML]{FF0000} \textit{\textbf{79.8711}}} & 0.0893 & {\color[HTML]{FF0000} \textit{\textbf{20.9724}}} & {\color[HTML]{FF0000} \textit{\textbf{7.5851}}} & {\color[HTML]{FF0000} \textit{\textbf{0.4637}}} & {\color[HTML]{FF0000} \textit{\textbf{0.4032}}} & 0.8146 & 0.0622 & \textit{\textbf{1.4421}} & {\color[HTML]{FF0000} \textit{\textbf{7.5851}}} & {\color[HTML]{FF0000} \textit{\textbf{9.0202}}} & {\color[HTML]{FF0000} \textit{\textbf{7.6387}}} & \textit{\textbf{1.0453}} \\
		\multirow{-4}{*}{Ours} & add+max & \textit{\textbf{79.1167}} & 0.0967 & \textit{\textbf{20.6914}} & \textit{\textbf{7.5819}} & \textit{\textbf{0.4593}} & \textit{\textbf{0.3968}} & 0.8117 & 0.0590 & 1.3709 & \textit{\textbf{7.5819}} & \textit{\textbf{8.9609}} & \textit{\textbf{7.5732}} & 1.0063 \\ \hline
	\end{tabular}
	}
\end{table*}


\section{CONCLUSIONS}
\label{section:CONCLUSIONS}
We developed a decomposition network fusion framework for fusing infrared and visible light images. 
It includes a novel multi-network for image decomposition. With the help of the decomposition network, 
the infrared image and the visible light image are decomposed into multiple high-frequency feature images and a low-frequency feature image, respectively. 
The corresponding feature maps are fused using a specific fusion strategy to obtain the fused feature maps. 
Finally, the fused band pass images reconstructed from the fused feature maps are added pixel by pixel to obtain the final fused image. 
The proposed image decomposition network is universal, and can be extended for use with any number of images.
In any case, the decomposition neural networks can use GPUs for matrix calculation acceleration. 

We performed subjective and objective evaluation of the proposed method. 
The experimental results show that the advocated method achieves the state of the art performance. 
The advantage of the proposed approach is the simplicity of the network structure. 
The subjective evaluation suggests that our CNN extracts the semantics of the image and filters out the noise, while preserving the edges, and other high-frequency information. 
We intend to study image decomposition based on deep learning, to simplify the complex image decomposition calculations such as wavelet transformation, low-rank decomposition, etc. 
We will also explore other network structures and address different applications. 
The network we proposed can be used for different image processing tasks, including multi-focus fusion, medical image fusion, multi-exposure fusion, and some basic computer vision tasks such as detection, recognition, and classification.

\ifCLASSOPTIONcaptionsoff
  \newpage
\fi
\bibliographystyle{IEEEtran}
\bibliography{IEEEabrv,paper}

\begin{IEEEbiography}[{\includegraphics[width=1in,height=1.25in,clip,keepaspectratio]{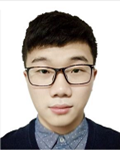}}]{Yu Fu}
	received the M.S. degree from Jiangnan University, China and A.B. degree from North China Institute Of Science And Technology, China. He is currently a Master student in the Jiangsu Provincial Engineerinig Laboratory of Pattern Recognition and Computational Intelligence, Jiangnan University. His research interests include image fusion, machine learning and deep learning.
\end{IEEEbiography}

\begin{IEEEbiography}[{\includegraphics[width=1in,height=1.25in,clip,keepaspectratio]{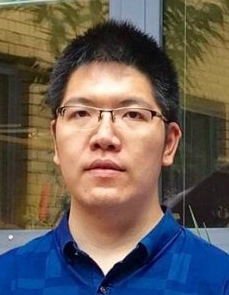}}]{Tianyang Xu} received the B.Sc. degree in electronic science and engineering from Nanjing University, Nanjing, China, in 2011. He received the PhD degree at the School of Artificial Intelligence and Computer Science, Jiangnan University, Wuxi, China, in 2019. 
He was a research fellow at the Centre for Vision, Speech and Signal Processing (CVSSP), University of Surrey, Guildford, United Kingdom, from 2019 to 2021. 
He is currently an Associate Professor at the School of Artificial Intelligence and Computer Science, Jiangnan University, Wuxi, China.
His research interests include visual tracking and deep learning. 
He has published several scientific papers, including IJCV, ICCV, TIP, TIFS, TKDE, TMM, TCSVT etc. He achieved top 1 tracking performance in several competitions, including the VOT2018 public dataset (ECCV18), VOT2020 RGBT challenge (ECCV20), and Anti-UAV challenge (CVPR20).
\end{IEEEbiography}

\begin{IEEEbiography}[{\includegraphics[width=1in,height=1.25in,clip,keepaspectratio]{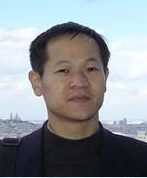}}]{Xiao-Jun Wu}
	received the B.Sc. degree in mathematics from Nanjing Normal University, Nanjing, China, in 1991, and the M.S. and Ph.D. degrees in pattern recognition and intelligent system from the Nanjing University of Science and Technology, Nanjing, in 1996 and 2002, respectively
	
	From 1996 to 2006, he taught at the School of Electronics and Information, Jiangsu University of Science and Technology, where he was promoted to a Professor. He was a Fellow of the International Institute for Software Technology, United Nations University, from 1999 to 2000. He was a Visiting Researcher with the Centre for Vision, Speech, and Signal Processing (CVSSP), University of Surrey, U.K., from 2003 to 2004. Since 2006, he has been with the School of Information Engineering, Jiangnan University, where he is currently a Professor of pattern recognition and computational intelligence. His current research interests include pattern recognition, computer vision, and computational intelligence. He has published over 300 articles in his fields of research. He was a recipient of the Most Outstanding Postgraduate Award from the Nanjing University of Science and Technology.
\end{IEEEbiography}

	

\vfill

\end{document}